\documentclass[10pt,twocolumn,letterpaper]{article}
\usepackage[accsupp]{axessibility}
\usepackage[pagenumbers]{iccv}
\usepackage{natbib}
\usepackage{xcolor}
\usepackage{pst-text}
\usepackage{amssymb}
\usepackage{pifont}
\usepackage{colortbl}
\usepackage{multirow}
\usepackage{mathrsfs}
\usepackage{titletoc}
\usepackage[most]{tcolorbox}
\usepackage[misc]{ifsym}

\newcommand{\cmark}{\ding{51}\xspace}%
\newcommand{\xmarkg}{\textcolor{lightgray}{\ding{55}}\xspace}%
\definecolor{myorange}{RGB}{255,100,3}
\definecolor{mygray}{gray}{.85}
\definecolor{mygray1}{gray}{.7}
\definecolor{mygray2}{gray}{.93}
\definecolor{mygray3}{gray}{.90}

\newcommand{\myparagraph}[1]{{\vspace{.1em} \noindent \bf #1}}

\definecolor{cvprblue}{rgb}{0.21,0.49,0.74}
\usepackage[pagebackref,breaklinks,colorlinks,allcolors=cvprblue]{hyperref}

\newcommand{\ourtask}{OmniAVS\xspace}

\newcommand{\ourdataset}{OmniAVS\xspace}
\newcommand{\ourmodel}{OISA\xspace}
\newcommand{\ourinstance}{OISA-1B\xspace}
\newcommand{\segtoken}{[SEG]\xspace}

\newcommand{\videonum}{2,104\xspace}

\newcommand{\objectnum}{4,277\xspace}

\newcommand{\explnum}{34,841\xspace}
\newcommand{\exprnum}{61,095\xspace}

\newcommand{\myrule}{\specialrule{.1em}{.05em}{.05em}}
\newcommand{\pub}[1]{\color{gray}{\scriptsize{[{#1}]}}}
\newcommand{\roundedcolorbox}[2]{%
  \begin{tikzpicture}[baseline=(char.base)]
    \node[fill=#1, rounded corners=5pt, inner sep=2pt] (char) {#2};
  \end{tikzpicture}%
}

\begin{document}

\title{
Towards Omnimodal Expressions and Reasoning in \\
Referring Audio-Visual Segmentation
}

\author{
Kaining Ying
\quad
Henghui Ding~\!$^{\textrm{\Letter}}$
\quad
Guangquan Jie
\quad
Yu-Gang Jiang
\vspace{.6mm}
\\
Fudan University, China
\vspace{.6mm}
\\
\href{https://henghuiding.com/OmniAVS/}{https://henghuiding.com/OmniAVS/}
}

\twocolumn[{
\renewcommand\twocolumn[1][]{#1}%
\maketitle
\begin{center}
    \centering
    \captionsetup{type=figure}
    \includegraphics[width=0.99\textwidth]{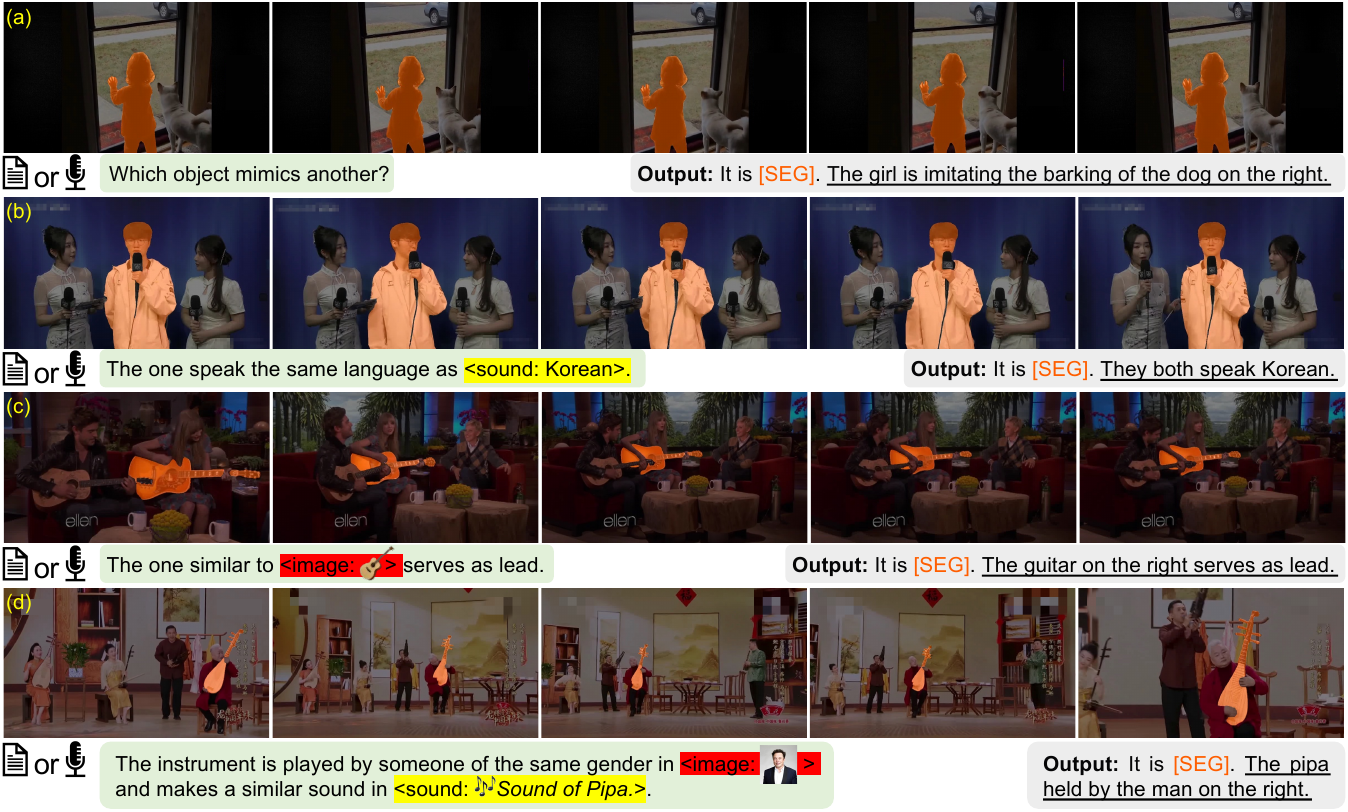}
    \vspace{-3.6mm}
    \captionof{figure}{\small 
    Examples of the proposed benchmark \textbf{Omni}modal Referring \textbf{A}udio-\textbf{V}isual \textbf{S}egmentation~(\textbf{\ourdataset}) to show its nature and flexibility.
    \ourtask introduces \textbf{3} key features: 
    \textbf{1)} It supports diverse multimodal referring expressions that flexibly combine \protect\roundedcolorbox{green!20}{text}, \protect\roundedcolorbox{green!10}{speech}, \protect\roundedcolorbox{yellow!70}{sound}, and \protect\roundedcolorbox{red!40}{image} for referring audio-visual segmentation;
    \textbf{2)} It emphasizes understanding the content of audio rather than merely hearing them;
    \textbf{3)} It incorporates complex reasoning and world knowledge in expressions, prompting models to provide \underline{explanations} for their segmentation decisions.
    These characteristics make \ourdataset practical for real-world use and well-suited for developing omnimodal models with fine-grained perception.
}
    \label{fig:teaser}
    \vspace{1.96pt}
\end{center}
}]
\renewcommand{\thefootnote}{\fnsymbol{footnote}}
\footnotetext[0]{${\textrm{\Letter}}$ Henghui Ding (henghui.ding@gmail.com) is the corresponding author with the Institute of Big Data, College of Computer Science and Artificial Intelligence, Fudan University, Shanghai, China.}

\begin{abstract}
Referring audio-visual segmentation (RAVS) has recently seen significant advancements, yet challenges remain in integrating multimodal information and deeply understanding and reasoning about audiovisual content.~To extend the boundaries of RAVS and facilitate future research in this field, we propose \textbf{Omni}modal Referring \textbf{A}udio-\textbf{V}isual \textbf{S}egmentation (\textbf{OmniAVS}), a new dataset containing \videonum videos and \exprnum multimodal referring expressions. OmniAVS stands out with three key innovations: (1) 8 types of multimodal expressions that flexibly combine text, speech, sound, and visual cues; (2) an emphasis on understanding audio content beyond just detecting their presence; and (3) the inclusion of complex reasoning and world knowledge in expressions.~Furthermore, we introduce \textbf{O}mnimodal \textbf{I}nstructed \textbf{S}egmentation \textbf{A}ssistant (\textbf{OISA}), to address the challenges of multimodal reasoning and fine-grained understanding of audiovisual content in OmniAVS. OISA uses MLLM to comprehend complex cues and perform reasoning-based segmentation. Extensive experiments show that OISA outperforms existing methods on OmniAVS and achieves competitive results on other related tasks.
\end{abstract}

  
\hyphenpenalty=1000
\tolerance=800
\vspace{-1mm}
\section{Introduction}
\label{sec:intro}
\vspace{-1mm}
Referring Audio-Visual Segmentation~(RAVS) \cite{refavs,ReferringSurvey} is an emerging field focused on segmenting target objects indicated by referring expressions (\eg, text) within audio-visual scenes using audio-visual cues.
The combination of \textit{referring} and \textit{audio} distinguishes RAVS from Referring Video Object Segmentation (RVOS)~\cite{mevis,refer_davis,refer_youtube_vos,primitivenet,losh,dshmp}, which uses text descriptions to refer to objects in \textit{silent} videos, and Audio-Visual Segmentation~(AVS)~\cite{avs,ovavss}, which segments any sound-producing object \textit{without} referring expressions.
RAVS enables practical applications in video conferencing and robotics, \etc.

\begin{figure}[t!]
    \centering
    \includegraphics[width=\columnwidth]{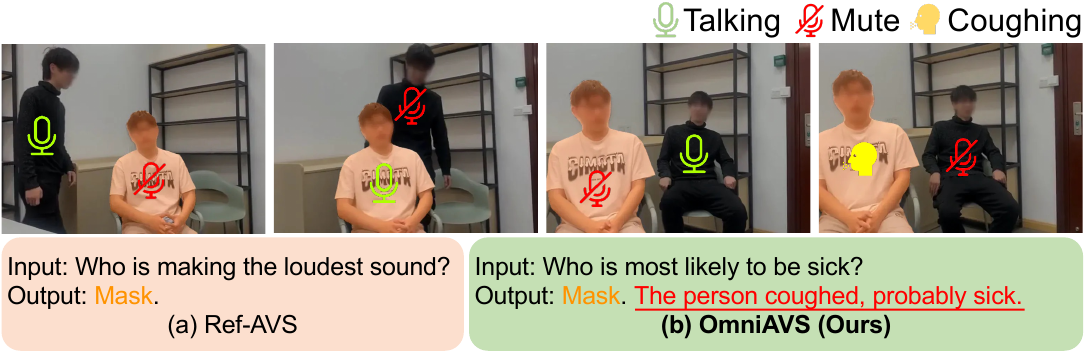}
    \vspace{-6mm}
    \caption{Comparison of (a) Ref-AVS~\cite{refavs} and (b) our proposed \ourdataset. In Ref-AVS, expressions mainly focus on surface-level sound properties, while \ourdataset demands deeper understanding and reasoning about sound content, enabling complex reasoning like identifying potential illness from coughing sounds.}
    \label{fig:compare_avs}
    \vspace{-1.6mm}
\end{figure}

\textbf{Understanding Rather than Just Hearing.}
However, the association between expressions and audio cues in the previous RAVS dataset Ref-AVS~\cite{refavs} remains superficial, primarily limited to basic acoustic properties such as sound occurrence, intensity, and temporal sequence. For example, expressions like ``\textit{Who is making the loudest sound?}" only address the surface-level characteristics without delving into the content of the sounds. 
Recent works~\cite{lisa,visa,videolisa,villa,detgpt} in image and video segmentation have evolved from simple semantic perception to advanced reasoning understanding. However, such progress is notably absent in audio-visual segmentation.
This gap naturally leads us to explore the potential for complex reasoning in audio-visual scenes. Besides, drawing insights from human auditory perception studies \cite{bregman1994auditory}, understanding sound content is essential for sophisticated audio-visual scene reasoning.
To achieve these goals, we present \ourdataset dataset to advance reasoning-based segmentation in audio-visual scenes.
As shown in \Cref{fig:compare_avs}(b), our dataset contains expressions that demand reasoning about audio content, such as ``\textit{Who is most likely to be sick?}". These expressions establish complex cognitive chains that go beyond basic acoustic features, \eg, sound$\rightarrow$coughing$\rightarrow$sickness. Furthermore, we enhance the dataset's interpretability by providing detailed explanations for reasoning-based expressions.

\textbf{Towards Omnimodal Referring.}~The emergence of flagship model ChatGPT-4o~\cite{gpt4o} sets a new standard for modern AI systems, emphasizing the need to process arbitrary combinations of omnimodal inputs, \eg, audio, vision, and text. 
In the context of RAVS, enabling omnimodal inputs can greatly enhance flexible referring and human-machine interactions. However, existing datasets \cite{mevis,refavs} are limited in scope, supporting only single or isolated modalities in their referring expressions.
To address this issue, we include diverse expressions in \ourdataset. Beyond the most common text-only expressions, \ourdataset supports text with sound, text with image, and text with both sound and image.
Additionally, we allow text to be replaced by speech\!\!~\footnote{To clarify, \textit{speech} is human-produced spoken sound.}, resulting in 8 different kinds of expression types in \ourdataset, as shown in \Cref{fig:teaser}.

\textbf{Omnimodal~Instructed~Segmentation~Assisant.}
Given the unique and complex requirements of \ourdataset, which necessitate both flexible interfaces and sophisticated multimodal perception and reasoning capabilities, no existing models can be directly applied to this emerging field. To address this issue, we introduce a baseline model named \textbf{O}mnimodal \textbf{I}nstructed \textbf{S}egmentation \textbf{A}ssistant, based on Multimodal Large Language Model (MLLM)~\cite{internvl}. This model seamlessly integrates text, speech, sound, and image inputs to perform referring object segmentation while providing explanations for prediction.
We identify audio-visual synchronization as a critical factor and implement Audio-Visual Interleaving to achieve temporal alignment without introducing additional parameters.
Furthermore, unlike concurrent works~\cite{videolisa} that use a single token to segment all frames, we adopt a more lightweight query propagation approach to achieve efficient and accurate segmentation.
Experimental results demonstrate that the proposed method~\ourmodel~possesses strong multimodal reasoning capabilities and supports multiple expression patterns, outperforming other methods on \ourdataset dataset. To further validate the effectiveness of our method \ourmodel, we conduct evaluations on other datasets, including referring video/image segmentation and Ref-AVS, achieving competitive results compared to existing methods across all tasks.

This work makes four main contributions: 
\textbf{i)} we explore omnimodal referring audio-visual segmentation, which is more aligned with real-world scenarios and provides a solid foundation for developing future omnimodal AI systems;
\textbf{ii)} we introduce a large-scale referring audio-visual segmentation dataset \ourdataset that enables advanced multimodal reasoning in audio-visual scenes, with a particular emphasis on deep understanding and reasoning from audio content;
\textbf{iii)} we enhance interaction flexibility by incorporating multiple referring modalities in the proposed \ourdataset dataset, broadening the scope of referring AVS; 
\textbf{iv)} we develop \ourmodel, a multimodal large language model designed to uniformly handle omnimodal expressions and reasoning in RAVS, with extensive experiments across different tasks.
\section{Related Work}
\label{sec:related_work}

\myparagraph{Audio-Visual Scene Perception.} Audio-visual scene perception \cite{music_avqa,avqa,refavs,hu_1,hu_2,hu_3,hu_4,hu_5,hu_6} aims to simulate human multimodal sensory capabilities by integrating auditory and visual information. 
The rapid development of deep learning has enabled joint audio-visual learning, achieving significant progress in tasks such as event detection \cite{eventlocalization}, robotic manipulation \cite{wang2022audio}, object localization \cite{objectlocalization}, action recognition \cite{actionrecognition}, and object segmentation \cite{avs,refavs}. 
The integration of multiple modalities enables models to achieve superior performance compared to single-modality approaches, particularly in handling incomplete or ambiguous information. However, datasets that comprehensively incorporate audio, visual, and language components remain scarce.
Among the few available datasets, Music-AVQA~\cite{music_avqa}, AVQA \cite{avqa}, and Ref-AVS \cite{refavs} stand out as notable examples.
Music-AVQA \cite{music_avqa} and AVQA \cite{avqa} primarily address question-answering tasks.
Ref-AVS \cite{refavs} focuses on referring audio-visual segmentation. Nevertheless, Ref-AVS's utilization of audio information is limited, as it employs relatively simple expressions without complex reasoning, thus constraining its effectiveness in real-world scenarios.
To overcome these limitations, we introduce \ourdataset, which is specifically designed to maximize the utilization of audio information and seamlessly integrate it with textual descriptions. 

\myparagraph{Reasoning-based Perception.}
With the rapid development of MLLMs~\cite{llava,llavav1.5,qwenvl,qwen2vl,qwen2audio,internvl,salmonn,videosalmonn,llama-adapter, mmtbench, convbench},
the research community has shifted from perception based on simple semantic~\cite{mevis,mutr} to perception based on world knowledge or complex reasoning~\cite{detgpt,lisa,videolisa,visa,gpt4roi,anyref}.
For example, GPT4RoI~\cite{gpt4roi} prompts MLLM with bounding boxes to solve region-level complex tasks that require reasoning.
DetGPT~\cite{detgpt} combines a pretrained MLLM with an open-vocabulary object detector to enable reasoning-based object detection from natural language instructions.
LISA~\cite{lisa} uses a MLLM~\cite{llava} to output a \segtoken token, which is then processed by a mask head~\cite{sam} to generate the masks.
Leveraging the world knowledge and reasoning capabilities inherent in LLMs, LISA demonstrates superior performance in complex segmentation scenarios that require complex reasoning.
Building upon this foundation, VideoLISA~\cite{videolisa} and VISA~\cite{visa} extend LISA's capabilities to reasoning video object segmentation by incorporating comprehensive temporal modeling.
Drawing inspiration from these works, we propose \ourmodel, which extends reasoning capabilities to audio-visual contexts. Unlike previous works that primarily focus on vision-language reasoning, \ourmodel leverages MLLM to perform reasoning about audio content in videos. While AnyRef~\cite{anyref} pioneers the use of audio and images as referring inputs for segmentation, it is constrained to static images and lacks the ability to reason about audio in dynamic video contexts.
Moreover, some works \cite{baichuan_omni,vita,gpt4o} extend LLMs to multiple modalities, \eg, vison, audio, and text, but they cannot be used for segmentation tasks.

\section{Benchmark: \ourdataset}
\label{sec:dataset}

\begin{table*}[t!]
    \centering
    \footnotesize
    \setlength\tabcolsep{4.2pt}
    \renewcommand\arraystretch{1.26}
    \caption{Statistical comparison between the newly proposed \ourdataset and other datasets of related referring video segmentation tasks. Avail., Expl., and Expr. are abbreviations for Availability of reasoning, Explanations, and Expressions, respectively.}
    \vspace{-3mm}
    \begin{tabular}{lc|cc|ccc|cc|cccccc}
        \myrule
        \rowcolor{mygray!30} &  & \multicolumn{2}{c|}{\textbf{Content}} & \multicolumn{3}{c|}{\textbf{Referring}} & \multicolumn{2}{c|}{\textbf{Reasoning}} & \multicolumn{6}{c}{\textbf{Statistics}} \\
         \rowcolor{mygray!30}\multirow{-2}{*}{\textbf{Dataset}}&\multirow{-2}{*}{\textbf{Venue}} & Video & Audio & Text & Audio & Image & Avail. & Expl. & Video & Frame & Object & Mask & Expr. & Expl. \\
        \hline
        Refer-YouTubeVOS~\cite{refer_youtube_vos} & \pub{ECCV'20} & \cmark & \xmarkg & \cmark & \xmarkg & \xmarkg & \xmarkg & \xmarkg & 3,978 & \textbf{116k} & 7,451 & 131k & 15,009 & \xmarkg \\
        \rowcolor{mygray2!36}MeViS~\cite{mevis} & \ \ \!\pub{ICCV'23} & \cmark & \xmarkg & \cmark & \xmarkg & \xmarkg & \xmarkg & \xmarkg & 2,006 & \ \ 44k & \textbf{8,175} & 443k & 28,570 & \xmarkg \\
        ReVOS~\cite{visa}  & \pub{ECCV'24} & \cmark & \xmarkg & \cmark & \xmarkg & \xmarkg & \cmark & \xmarkg & 1,042 & 116k & 5,535 & \textbf{469k} & 35,074 & \xmarkg \\
        \rowcolor{mygray2!36}Ref-AVS~\cite{refavs}& \pub{ECCV'24} & \cmark & \cmark & \cmark & \xmarkg & \xmarkg & \xmarkg & \xmarkg & \textbf{4,002} & \ \ 40k & 6,888 & \ \ 78k & 20,261 & \xmarkg \\
        \hline
        \rowcolor{cyan!10} 
        \textbf{\ourdataset (ours)}& \pub{ICCV'25} & \cmark & \cmark & \cmark & \cmark & \cmark & \cmark & \cmark & \videonum & 103k & \objectnum & 206k & \textbf{\exprnum} & \textbf{\explnum} \\
        \specialrule{.1em}{.05em}{.05em}
    \end{tabular}
    \label{tab:dataset_comparison}
\end{table*}

\subsection{Audio-Visual Video Collection}
\label{sec:video_collection}
The video sources consist of three main parts: 
i) Real-world web videos under the \textit{Creative Commons License}, where we use techniques from \cite{avs,refavs,vggsound} to align audio and visual snippets with intended semantics; 
ii) The open-source dataset TVQA~\cite{tvqa}, featuring extensive dialogue from TV shows; 
iii) Self-recorded videos, with consent from all participants. 
We select videos from above that meet the following two standards:

\begin{itemize}
\item We encourage to select videos that contain audio content with informative or reasoning-based meanings and corresponding identifiable objects in the scene.
\item We strive to select videos with complex scenes and multiple objects, enabling more diverse referring expressions, \eg, referring to multiple targets, and makes the dataset more representative of real-world complex scenarios.
\end{itemize}

After reviewing about 10,871 potential candidates, we carefully selected the most satisfactory and appropriate videos, which meet the above standards. Finally, we chose \videonum videos to create the \ourdataset dataset, which is diverse and representative of a wide range of real-world audio-visual scenarios. 
After collecting the videos, our research team annotates potential objects with expressions and bounding boxes at their first appearance.
This information is then used by the annotation team to achieve video mask annotation on the referred objects.
Next, we will introduce the corresponding annotation process.

\subsection{Omnimodal Expression and Mask Annotation}
\label{sec:anno}
\begin{figure}[t!]
    \centering
    \begin{minipage}{0.51\linewidth}
        \centering
        \captionof{table}{\centering More dataset statistics.}
        \label{tab:dataset_card}
        \vspace{-3mm}
        \footnotesize
        \setlength\tabcolsep{1.4pt}
        \begin{tabular}{l|l}
            \specialrule{.1em}{.05em}{.05em}
            \rowcolor{mygray!30}Statistic & Number \\
            \hline
            Words per Expr. & 9.3 \\
            \rowcolor{mygray2!36}FPS (mean/min/max) & 4.8/3/15 \\
            Expression Types & 8 \\
            \rowcolor{mygray2!36}Frames (min/max) & 11/251     \\
            Obj. per Expr. (mean)  &  1.12              \\
            \rowcolor{mygray2!36} Obj. per Expr. (max)  &    7              \\
            Duration & 6.2 h \\
            \rowcolor{mygray2!36}Videos (train/test)  &    1,864/240              \\
            Expr. (train/test)  &     54,304/6,791 \\
            \specialrule{.1em}{.05em}{.05em}
        \end{tabular}
    \end{minipage}
    \hfill
    \begin{minipage}{0.45\linewidth} 
        \centering
        \caption{\centering Distribution of expression types. \text{SI}: \textit{sound+image}.}
        \label{fig:expr_distribution}
        \vspace{-3mm}
        \includegraphics[width=0.70\linewidth]{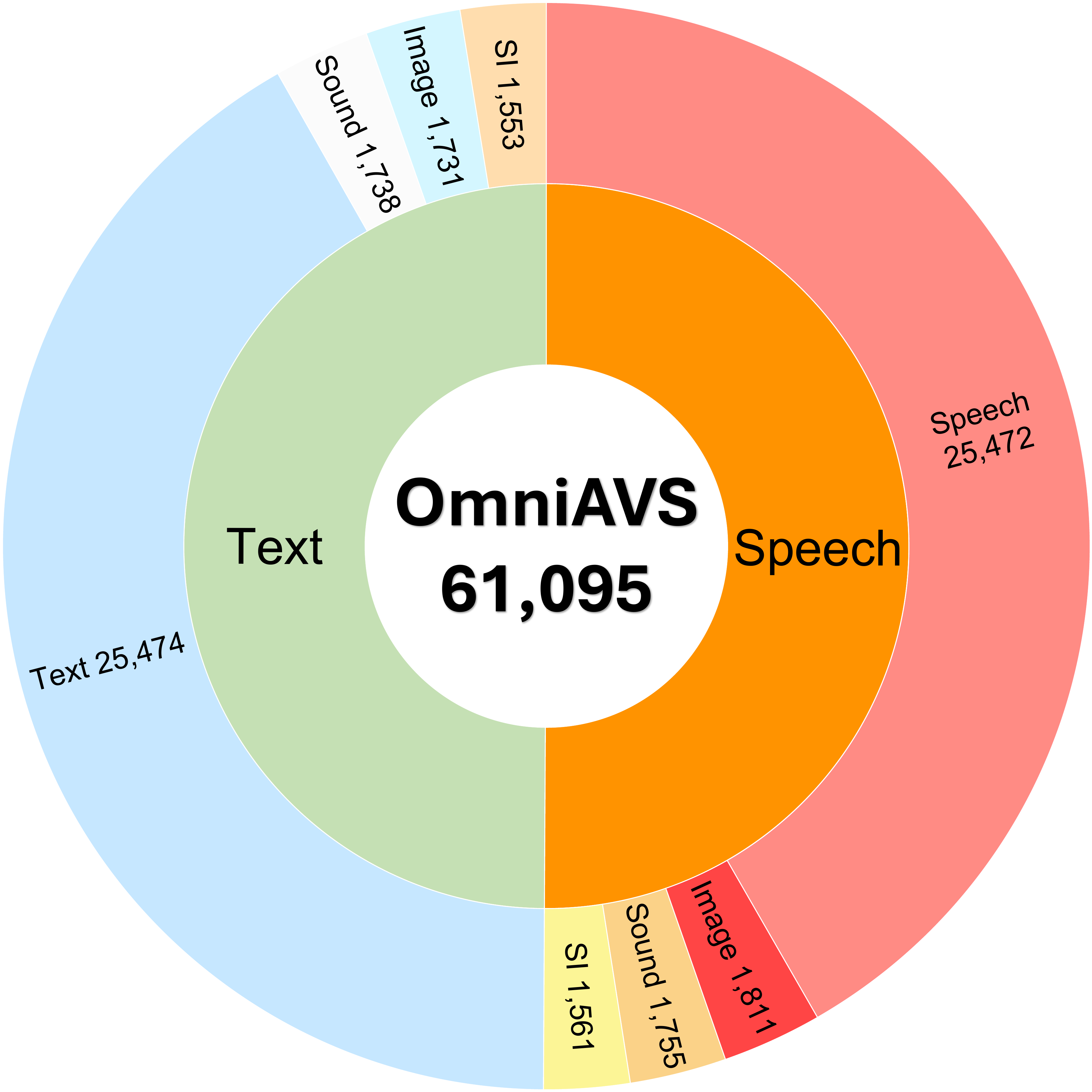}
        
    \end{minipage}
    \vspace{-3mm} 
\end{figure}

The expression annotation process integrates 4 different modalities: \textit{text}, \textit{speech}, \textit{sound}, and \textit{image}. 
We categorize referring expressions into 8 forms: 
\textbf{I)} Text;
\textbf{II)} Speech;
\textbf{III)} Text with Sound;
\textbf{IV)} Speech with Sound;
\textbf{V)} Text with Image;
\textbf{VI)} Speech with Image;
\textbf{VII)} Text with Sound and Image; and
\textbf{VIII)} Speech with Sound and Image.
Each expression includes either text or speech, as they provide essential instructions for the model to understand the task and other modalities. All speech expressions are generated by converting the corresponding text expressions into audio. Following previous works~\cite{wnet,vita}, we use a combination of several TTS models~\cite{openai_tts,gpt_sovits} and human readings for this conversion.
In light of this, we will focus on introducing the annotation process for text-related expressions. \Cref{fig:expr_distribution} shows the distribution of different expression types. 
The inner areas~(\eg, text) are composed of their corresponding outer areas~(\eg, 4 components of text).

\myparagraph{I) Text.}
This is the main component of \ourdataset. The annotation process follows these rules:
1) Expressions should relate to the sound in the video, not just visual cues.
2) Expressions should emphasize the sound's content, not just the act of making it. For example, instead of ``\textit{The dog barking}", use ``\textit{The dog warning}" as it requires understanding the sound.
3) Encourage reasoning in expressions, with necessary explanations.
4) Expressions can refer to any number of objects, from none to many (more than one).

\myparagraph{III) Text with Sound.}~This form combines text and sound to provide a more comprehensive expression. The annotation process 
follows the same rules as in the \textit{Text} modality, with the following additional requirements when 
adding audio:
1) Use natural sounds, instrumental music, or environmental sounds, \eg, dog barking, bird chirping.
2) Text should complement the sound, adding context or instructions for identifying the object(s). An example is shown in the 2nd row of \Cref{fig:teaser}.

\myparagraph{V)) Text with Image.}
This type combines text and image. The annotation follows the rules of \textit{Text} modality with additional considerations for images:
1) Include relevant images that depict the target object(s) or provide context for the audio-visual content.
2) Ensure the text complements the image, offering instructions for identifying the object(s). An example is shown in the 3rd row of \Cref{fig:teaser}.

\myparagraph{VII) Text with Sound and Image.}
This type combines text, sound, and image in the expressions. The text should guide identification using both sound and image.
An example is shown in the 4th row of \Cref{fig:teaser}.

\textbf{Mask Annotation.}~After collecting expressions, the annotation team proceeds to segment the objects referred to by each expression in every video. Using the bounding box of each object's first appearance (not necessarily in the first frame) annotated in the previous step as a reference, the annotation team is required to carefully confirm the corresponding object in the video, then track and annotate the object's mask in all frames
\footnote{We follow MeViS~\cite{mevis} to perform full-temporal segmentation, even when objects only partially match expressions in specific time segments.}.
We designed an interactive annotation tool to automatically load the video and corresponding object bounding boxes. Annotators use the tool to load and preview the corresponding video, then sequentially track and segment target objects in the video. The tool integrates state-of-the-art interactive video object segmentation model SAM2~\cite{sam2} to assist with mask annotation, enhancing efficiency.

\subsection{Dataset Statistics and Analysis}
\label{sec:data_analysis}

\Cref{tab:dataset_comparison} shows a statistical comparison between \ourdataset and related datasets.~Refer-YouTubeVOS~\cite{refer_youtube_vos}, MeViS~\cite{mevis}, and ReVOS~\cite{visa} focus on \textit{silent} videos without audio and provide only text expressions.~While ReVOS~\cite{visa} further supports reasoning expression, it lacks explanations. Ref-AVS~Bench~\cite{refavs} includes audio-visual videos but supports only single-modality text expressions without reasoning or explanations. Compared to previous datasets, the proposed \ourdataset dataset offers multimodal content (audio-visual videos) and diverse expression modalities (text, speech, sound, image), supports reasoning with explanations, and allows for referring to arbitrary number of target objects. We compare more datasets in supplementary.

\textbf{Main Differences with Ref-AVS.}
As Ref-AVS~\cite{refavs} is the most related dataset to our proposed \ourdataset, we summarize the differences here to facilitate reference:
\begin{itemize}
    \item \textbf{Omnimodal Expressions}: While Ref-AVS uses only textual referring expressions, \ourdataset supports expressions that can combine text, speech, sound, image, providing a richer multimodal interactive interface.

    \item \textbf{Enhanced Audio-Visual Integration}: Unlike Ref-AVS, which primarily focuses on ``the sounding object", \ourdataset offers a more comprehensive audio-visual analysis. Our expressions delve deeper into the content of sounds and their contextual impact on the scene.

    \item \textbf{Reasoning and Explanation}: Our dataset includes numerous expressions requiring reasoning, such as ``\textit{Who is most likely to be sick?}'' (see~\Cref{fig:compare_avs}), which necessitates inferring from audio-visual cues like sneezing. We also provide explanations for some reasoning expressions.
    
    \item \textbf{Number of Referred Target Objects}: Unlike Ref-AVS that only refers to 0 or 1 object per expression, \ourdataset allows each referring expression to indicate an arbitrary number of target objects, \ie, from 0 to many as \cite{gres}. 
    
    \item \textbf{Annotation FPS}: While Ref-AVS uses fixed video settings (10s, 1 FPS), \ourdataset includes videos of varying durations and annotation frame rates (3-15 FPS) to capture temporal dynamics. \Cref{tab:dataset_card} provides more statistics.
\end{itemize}

\section{Method: \ourmodel}
\label{sec:method}

\subsection{Overview Architecture}
\label{sec:overview}
\begin{figure}[t]
    \centering
    \includegraphics[width=\columnwidth]{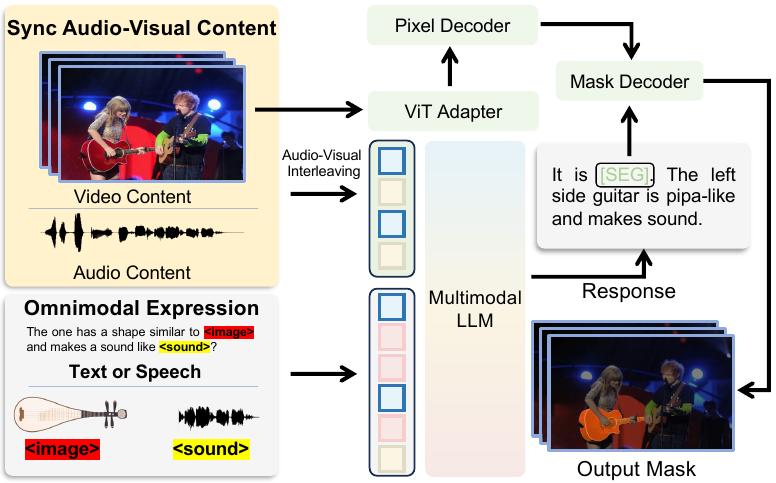}
    \vspace{-6mm}
    \caption{{Overview of the proposed method \textbf{\ourmodel}.} 
    Vision/audio encoders and text embedding are omitted for clarity. }
    \label{fig:framework}
    \vspace{-1.6mm}
\end{figure}

Because of the unique nature of this task and the proposed \ourdataset dataset, there is no off-the-shelf model available. We propose an Omnimodal Instructed Segmentation Assistant~(\textbf{\ourmodel}) as a baseline method to support omnimodal referring and reasoning audio-visual segmentation.~As shown in \Cref{fig:framework}, \ourmodel consists of two main components: 
a MLLM for multimodal understanding and generation, and a mask head for segmentation and tracking.
The MLLM includes audio encoder, vision encoder and LLM, while the mask head consists of ViT-Adapter~\cite{vitadapter}, pixel decoder and mask decoder~\cite{mask2former}.
It's worth noting that the mask head is flexible and can be replaced with models like SAM~\cite{sam}. 

Given an audio-visual video, the multimodal encoder processes video and audio content to obtain corresponding vision and audio tokens. 
Different from VideoLLaMA~\cite{videollama} that concatenates these two modalities sequentially, we adopt an Audio-Visual Interleaving strategy, as described in \Cref{sec:audio_visual_interleaving}, where we divide audio tokens into clips and interleave them with vision tokens. The interleaved sequence forms our \textit{Audio-Visual Content Tokens}.
This approach effectively synchronizes audio and frames without introducing additional parameters, 
which is particularly useful for scenarios requiring audio-visual alignment.
See examples at \Cref{fig:compare_avs}, where without proper audiovisual alignment it would be difficult to determine which object sneezed when multiple objects are making sounds simultaneously.
For omnimodal expression inputs, we employ the audio encoder to process speech and sound, and the vision encoder to handle images. The resulting tokens are then integrated into the corresponding text tokens to produce the final \textit{Omnimodal Expression Tokens}. All types of tokens are subsequently fed into the MLLM. The MLLM generates text responses and produces a \segtoken token representing the target object, which is then used by mask decoder for segmentation prediction.
Our mask decoder uses query propagation to segment each frame, as described in \Cref{sec:query_propagation}. The query is refined online during the segmentation process, alleviating dynamic motion.
Following previous work~\cite{lisa}, we employed cross entropy loss for text generation, DICE loss~\cite{diceloss}, and binary cross entropy loss for segmentation. 

\subsection{Audio-Visual Interleaving}
\label{sec:audio_visual_interleaving}
For videos of arbitrary length, we uniformly sample \( N \) frames from the beginning to the end. Each frame is individually processed by the vision encoder to obtain \( L_v \) vision tokens, denoted as
\( \mathbf{V} = \{v_1, v_2, \dots, v_N\} \), where \( v_i \in \mathbb{R}^{L_v \times d} \) represents the \( L_v \) vision tokens of the \( i \)-th frame with dimension \( d \).
For the audio content in the video, we use an audio encoder to process it and obtain audio tokens \( \mathbf{A} \in \mathbb{R}^{L_A \times d} \), where \( L_A \) denotes the total number of audio tokens. 
To achieve audio-visual alignment, we segment the audio into clips corresponding to the frame rate, obtaining \( \{a_1, a_2, \dots, a_N\} \), where \( a_i \in \mathbb{R}^{L_a \times d} \) and \( L_a = L_A / N \).
Next, we interleave audio token clips and vision tokens to form the audio-visual interleaving token sequence \( \{v_1, a_1, v_2, a_2, \dots, v_N, a_N\} \). 
In comparison, VideoLLaMA~\cite{videollama,videollama2} directly concatenates audio and visual tokens sequentially, \( \{v_1, \dots, v_N, a_1, \dots, a_N\} \). However, this approach makes it difficult to align audio frames, especially in segmentation tasks where the video's FPS is not fixed, thereby exacerbating the challenge of audio-visual synchronization.
Besides, video-SALMONN~\cite{videosalmonn} fuses audio and visual tokens by concatenating them along the feature dimension and then applying a weight matrix for mapping. This way introduces additional training parameters and causes the fused audio-visual features becoming misaligned with the original pretrained MLLM model.

\subsection{Query Propagation}
\label{sec:query_propagation}
\begin{figure}[t!]
    \centering
    \includegraphics[width=0.99\columnwidth]{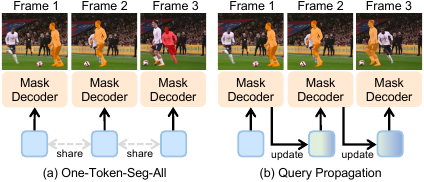}
    \vspace{-1.6mm}
    \caption{Comparision of different mask decoder forms.}
    \label{fig:mask_head}
    \vspace{-1mm}
\end{figure}

As shown in \Cref{fig:framework}, once the Audio-Visual Content Tokens and Omnimodal Expression Tokens are obtained, the MLLM generates the corresponding responses. We employ the \segtoken token, following LISA~\cite{lisa}, to represent the object embedding, which is subsequently passed to the mask decoder.
While MLLMs excel at multimodal understanding and reasoning, they are not good at segmentation tasks. Previous works~\cite{lisa,videolisa} attach an additional vision encoder for plain feature extraction and a mask head for segmentation after MLLM, resulting in a redundant and suboptimal design.
To address these issues, when extracting vision tokens from video, we simultaneously extract corresponding multi-scale features through ViT Adapter~\cite{vitadapter}, which are then enhanced via pixel decoder~\cite{mask2former}.
These enhanced features, along with the \segtoken generated by the MLLM, are input into the mask decoder.
When using mask decoder for segmentation, VideoLISA~\cite{videolisa} employs the same \segtoken for the independent segmentation of each frame, \textit{aka} One-Token-Seg-ALL~(OTSA), as shown in \Cref{fig:mask_head}~(a). However, this approach has limitations.
Previous studies \cite{vita,dvis,genvis,ctvis,isda} have shown that a single query often fails to adequately represent a target object, especially when rapid motion is present within the video. A single query carries a positional prior, making it difficult to encapsulate dynamic motion processes, such as an object moving from right to left, as shown in \Cref{fig:mask_head}~(a). This limitation results in ID switches, {consistently tracking the target on the right part of the video}.
To address this, we employ a query propagation mechanism, as shown in \Cref{fig:mask_head}~(b), which allows the query to be updated online on a frame-by-frame basis~\cite{genvis,motr}. This approach enables the query to smoothly capture the temporal motion of the object while also effectively modeling contextual temporal information.

\section{Experiment}
\label{sec:experiment}

\textbf{Implementation Details.}
We adopt InternVL2-1B~\cite{internvl}~as our MLLM, equipped with InternViT-300M-448px~\cite{internvl} as vision encoder and Qwen2-0.5B-Instruct~\cite{qwen} as LLM. Whisper-large-v3~\cite{whisper} is used as our audio encoder, with an extra audio MLP to project audio features into LLM feature space. For ViT-Adapter~\cite{vitadapter}, we remove all Injectors to reduce memory consumption and ensure vision tokens remain unaffected. For pixel decoder and mask decoder, we follow the default design of Mask2Former~\cite{mask2former} but remove the self-attention module in each transformer block since there is only one query input to the mask decoder.
We denote our model with the above settings as \ourinstance.

\textbf{Dataset Split.}
We split \ourdataset into training and testing sets, containing 1,864 videos with 54,304 expressions and 240 videos with 6,791 expressions, respectively. 

\textbf{Evaluation Metrics.} 
Following previous work~\cite{refavs,mose,MOSEv2,mevis,MOVE},
we adopt $\mathcal{J}\&\mathcal{F}$ as the segmentation evaluation metric, where $\mathcal{J}$ measures region similarity (IoU) and $\mathcal{F}$ evaluates contour accuracy.
Following \cite{gres}, for no-target expressions that referring to nothing in the video, $\mathcal{J}\&\mathcal{F}$ is set to 1 if the prediction is empty; otherwise, it is 0. We employ METEOR~\cite{meteor} as evaluation metric for text explanation generation.

\subsection{Training Details}
\label{sec:train_inference}
\textbf{Audio-Text Alignment.} 
Since InternVL2 does not support audio input natively, we introduce an audio encoder MLP to align audio features with the LLM space. 
This stage mainly aims to align audio features with LLM space. To achieve this goal, we use Automatic Speech Recognition datasets~\cite{gigaspeech} and Audio Caption datasets~\cite{autoacd} for audio-text alignment.
During this process, we only train the audio encoder MLP while keeping all other parameters frozen.

\textbf{Omnimodal Instruct~Segmentation Tuning.}~Following previous works~\cite{lisa,villa,visa,videolisa}, we use multiple datasets for training, including~1) semantic segmentation datasets: ADE20K~\!\cite{ade20k}, COCO-Stuff~\!\cite{cocostuff}, PASCAL-Part \cite{pascalpart}, PACO-LVIS~\!\cite{pacolvis}; 2) referring segmentation datasets: RefCOCO, RefCOCO+ \cite{Yu}, RefCOCOg \cite{refcocog}, ReasonSeg \cite{lisa}; 4) referring video segmentation datasets: Refer-YouTube-VOS \cite{refer_youtube_vos}, Refer-DAVIS-17 \cite{refer_davis}, MeViS \cite{mevis}, ReVOS \cite{visa}; 5) audio-visual segmentation datasets: Ref-AVS Bench~\cite{refavs}, \ourdataset.
We use LoRA~\cite{lora} to fine-tune LLM, train all parameters of the mask head, and freeze all remaining parameters. 
For video samples, during training, we uniformly sample 10 frames from the entire video and follow~\cite{videolisa} to use 4 frames as dense frames and the remaining frames as sparse frames. During inference, we sample 32 frames with 4 frames as dense frames.

\subsection{Ablation Study}

\begin{table}[t!]
    \footnotesize
    \centering
    \caption{Ablation study on \textbf{audio-visual fusion}.~AVI~means Audio-Visual Interleaving.~TVQA Split represents the performance evaluated on samples collected from TVQA~\cite{tvqa}. These samples demand precise audio-visual alignment and feature a substantial amount of dialogue. See an example in \Cref{fig:qualitative}~(a).
    }
    \label{tab:audio_visual_fusion}
    \renewcommand{\arraystretch}{1.1} 
    \vspace{-3mm}
    \setlength\tabcolsep{10pt}
    \resizebox{0.47\textwidth}{!}{
    \begin{tabular}{lccc}
    \toprule
    \textbf{Fusion Type} & \textbf{Sketch} & \textbf{TVQA Split} & \textbf{Overall} \\
    \myrule
    Attention & - & 37.4 & 35.8 \\
    Concat & \raisebox{-0.3\height}{\includegraphics[height=0.3cm]{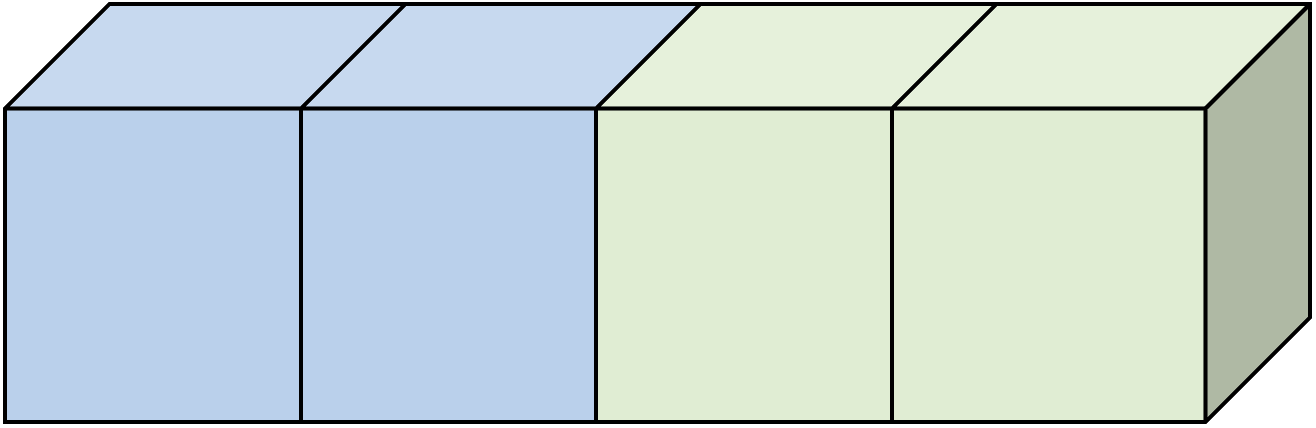}} & 36.9 & 35.3 \\
    Weighted Sum & \raisebox{-0.3\height}{\includegraphics[height=0.3cm]{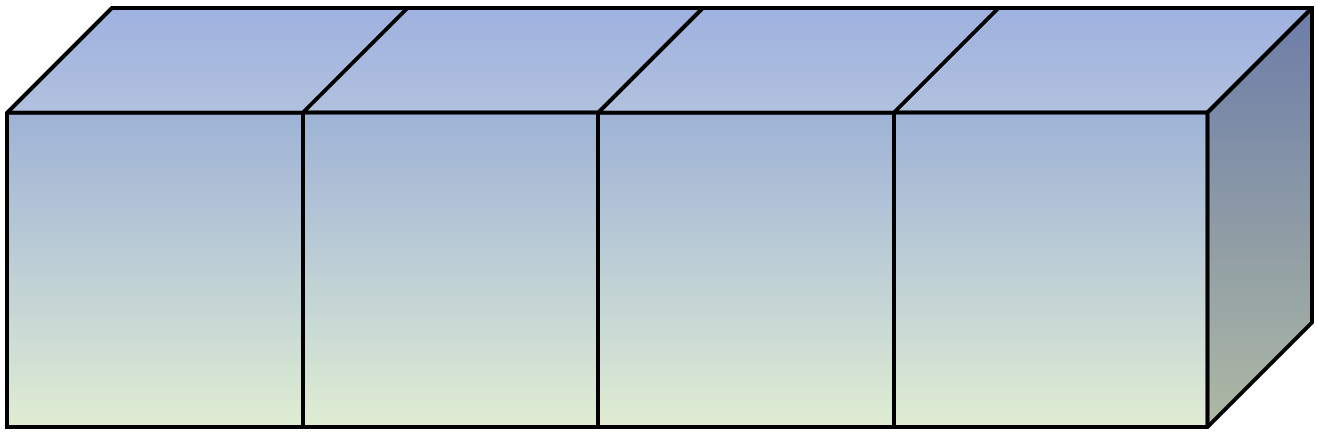}} & 34.8 & 31.1\\
    AVI & \raisebox{-0.3\height}{\includegraphics[height=0.3cm]{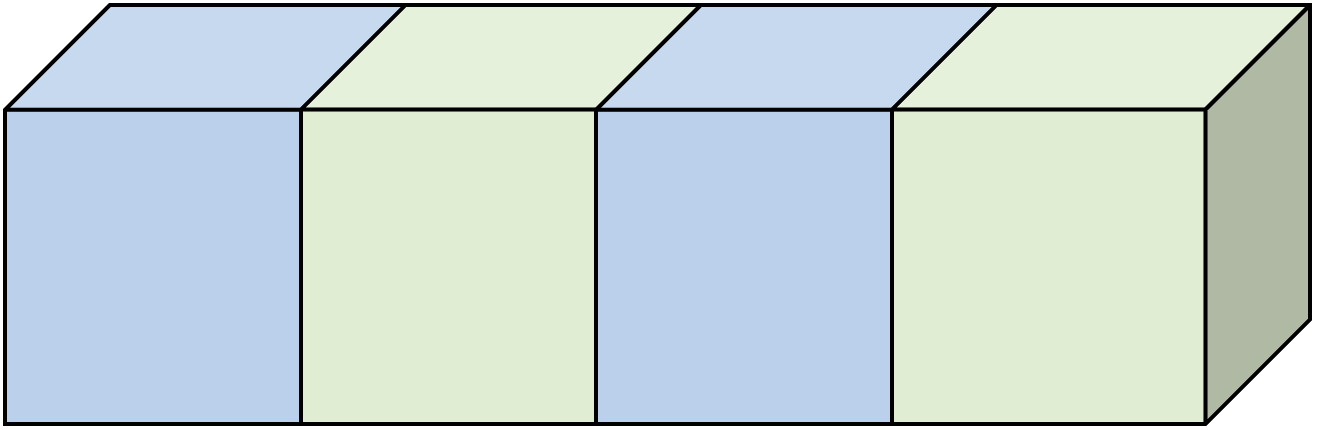}} & 40.8 & 39.2 \\
    AVI + Concat & \raisebox{-0.3\height}{\includegraphics[height=0.3cm]{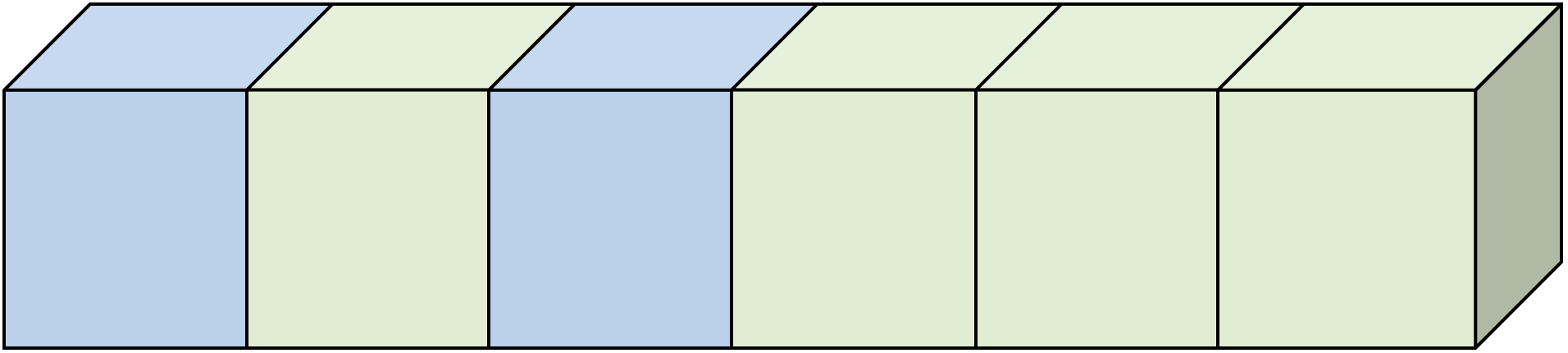}}  & 42.0  & 40.5 \\
    \myrule
    \end{tabular}%
    \renewcommand{\arraystretch}{1.}
    }
    \vspace{-0.6mm}
\end{table}
\begin{table}[ht]
    \centering
    \caption{Ablation study of \textbf{mask head}. OTSA: One-Token-Seg-All~\cite{videolisa}; QP: query propagation~(Ours); M2F: Mask2Former. FPS is tested on one A6000 GPU (batch size=1).}
    \vspace{-3mm}
    \footnotesize
    \setlength\tabcolsep{5pt}
    \label{tab:mask_head}
    \resizebox{0.476\textwidth}{!}{
    \begin{tabular}{l cc cc cccc}
    \toprule
    \multirow{2}*{\textbf{ID}} & \multicolumn{2}{c}{\textbf{Query Type}} & \multicolumn{2}{c}{\textbf{Head Type}} & \multicolumn{4}{c}{\textbf{\ourdataset} Metrics} \\
    ~ & \multicolumn{1}{c}{OTSA} & \multicolumn{1}{c}{QP} & \multicolumn{1}{c}{M2F} & \multicolumn{1}{c}{SAM} & \multicolumn{1}{c}{$\mathcal{J}\&\mathcal{F}$} & \multicolumn{1}{c}{$\mathcal{J}$} & $\mathcal{F}$ &  \multicolumn{1}{c}{FPS} \\
    \myrule
    I   & \cmark   & \xmarkg & \xmarkg & \cmark   & 38.1 & 36.4 & 39.8 & \ \ 4.3 \\
    II  & \cmark   & \xmarkg & \cmark  & \xmarkg  & 35.2 & 33.8 & 36.6 & 15.7 \\
    III & \xmarkg  & \cmark  & \xmarkg & \cmark   & 41.2 & 39.2 & 43.2 & \ \ 4.1 \\
    IV  & \xmarkg  & \cmark  & \cmark  & \xmarkg  & 40.5 & 38.7 & 42.3 & 12.3 \\
    \bottomrule
    \end{tabular}%
    }
    \vspace{-3mm}
\end{table}

\textbf{Ablation Study on Audio-Visual Fusion.}
In \Cref{tab:audio_visual_fusion}~we evaluate different audio-visual fusion strategies. For attention baseline, we adopt a simple cross-attention block where image tokens serve as key and audio tokens serve as query/value. For weighted sum, we first pad the audio tokens to match the length of vision tokens, then perform weighted sum on them. Attention-based and concatenation~\cite{videollama2} methods achieve similar performance around 35-37\% $\mathcal{J}\&\mathcal{F}$. 
Our proposed Audio-Visual Interleaving (AVI) strategy significantly improves the overall performance to 39.2\% and get 40.2\% in TVQA split.
Previous works~\cite{internvl,llavav1.5} add a thumbnail after tiling to capture global multimodal features to capture global features more effectively. Similarly, we append original audio tokens $\mathbf{A}$ at the end of the audio-visual interleaving sequence. 
This supplements complete and untruncated audio tokens based on our AVI audio-visual alignment, enabling the MLLM to handle more comprehensive audio tokens and thereby improving the model performance to 40.5\% $\mathcal{J}\&\mathcal{F}$.

\textbf{Ablation Study on Mask Head.} 
In \Cref{tab:mask_head} we study different mask head designs. While SAM mask head achieves competitive performance compared to M2F, it suffers from slower inference speed due to its huge ViT backbone. When using M2F as the mask head (Setting II and IV), we achieve at least 3$\times$ speedup with a slight accuracy trade-off. Regarding query types, our QP approach (Setting III and IV) significantly outperforms OTSA in segmentation quality, improving $\mathcal{J}\&\mathcal{F}$ from 38.1\% to 41.2\% when using SAM mask head, and from 35.2\% to 40.5\% when using M2F mask head. Considering the balance between performance and efficiency, we choose Setting IV (QP with M2F head) for our subsequent experiments.

\begin{table}[ht]
    \footnotesize
    \setlength\tabcolsep{1.86pt}
    \centering
    \caption{Testing on \textbf{\ourdataset}. 
    We use $\mathcal{J}\&\mathcal{F}$ as the default metric.
    \textit{All} is the average result across 8 splits. 
    \textit{MET.}: METEOR.
    }
    \vspace{-3mm}
    \label{tab:omniavs_result}
    \resizebox{0.476\textwidth}{!}{
        \begin{tabular}{lccccccccc|c}
            \toprule
            \textbf{Method}      & \textbf{All} & \textbf{I}    & \textbf{II}   & \textbf{III}    & \textbf{IV}     & \textbf{V}      & \textbf{VI}     & \textbf{VII}        & \textbf{VIII}  & \textbf{\textit{MET.}}       \\
            \myrule
            LMPM~\cite{mevis}     & 25.8 & 31.2 & 28.7 & 20.0 & 22.7 & 21.3 & 20.9 & 30.0 & 31.4 & -    \\
            EEMC~\cite{refavs}     & 29.6 & 34.4 & 32.6 & 19.6 & 26.0 & 28.0 & 24.7 & 35.6 & 36.0 & -    \\
            MUTR~\cite{mutr}     & 32.3 & 35.4 & 33.3 & 28.4 & 29.8 & 26.5 & 22.8 & 41.6 & 40.5 & -    \\
            LISA-7B~\cite{lisa}  & 33.6 & 33.3 & 31.2 & 29.2 & 32.7 & 28.6 & 27.3 & 43.4 & 43.1 & 11.6 \\
            LISA-13B~\cite{lisa} & 36.1 & 36.4 & 32.1 & 30.4 & 35.7 & 31.6 & 30.2 & 46.7 & 45.7 & 16.5 \\
            \rowcolor{cyan!10} 
            \textbf{\ourinstance} (\textbf{ours}) & \textbf{41.1} & \textbf{40.1} & \textbf{38.5} & \textbf{34.9} & \textbf{38.5} & \textbf{35.9} & \textbf{35.2} & \textbf{52.6} & \textbf{53.0} & \textbf{21.7} \\ 
            \myrule
        \end{tabular}%
    }
    \vspace{-1mm}
\end{table}

\subsection{\textbf{\ourdataset} Benchmark Results}
In \Cref{tab:omniavs_result}, we report the \textbf{\ourdataset} benchmark results.
For traditional referring segmentation methods~\cite{refavs,mevis,mutr}, we process each modality independently and concatenate their features to create a unified query representation. 
Regarding LISA~\cite{lisa}, we enhance its audio processing capability by applying the same audio-text alignment method as our \ourmodel, while maintaining its frame-by-frame inference approach.
Our \ourinstance achieves state-of-the-art performance on the \ourdataset benchmark, significantly surpassing previous methods. Specifically, it attains an average score of 41.1\% across all modality combinations, outperforming LISA-13B by 5.0\%. These improvements are consistent across all splits, with the most notable gain of 7.3\% on split VIII (speech+sound+image), showcasing our model’s superior ability in multimodal input processing and reasoning. In terms of explanation generation, our model achieves a METEOR score of 21.7\%, significantly surpassing LISA-13B’s score of 16.5\%. This indicates that our model excels not only in segmentation tasks but also in understanding and explaining its decisions. Analysis across different modality combinations reveals that splits VII (text+speech+image) and VIII (text+sound+image) achieve the highest scores of 52.6\% and 53.0\%, respectively. This suggests that incorporating more modalities generally enhances performance, as the model effectively leverages complementary information from various sources. Comparison with the previous dataset Ref-AVS Bench shows that our OISA-1B achieved a score of 58.0\% (shown in \Cref{tab:refavs}), whereas it only attained 41.1\% on our \ourdataset. This significant difference underscores the increased difficulty and complexity of our dataset.

\begin{figure}[t]
    \centering
    \includegraphics[width=.99\linewidth]{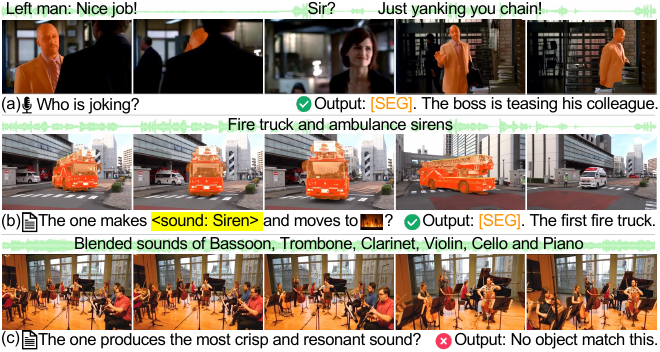}
    \vspace{-2mm}
    \caption{Success and failure cases of \ourinstance.}
    \label{fig:qualitative}
\end{figure}

\textbf{Qualitative Results and Failure Cases.} In \Cref{fig:qualitative}, we present several success and failure cases of the proposed baseline method \ourmodel.
Case~(a) demonstrates that \ourmodel effectively aligns audio with visual content, identifies key dialogues in the fourth frame, and subsequently performs reasoning and accurately locates the target object, providing a clear explanation.
Besides, Case~(a) shows that \ourmodel achieves robust tracking in scenarios involving viewpoint changes, occlusions, and reappearances.
Case~(b) illustrates the model's ability to handle multimodal expressions by understanding both the fire in the image and the siren sound to locate the fire truck.
However, as shown in Case~(c), \ourmodel struggles in scenarios with complex sounds, indicating that understanding intricate sounds still requires further exploration, such as decoupling sound events~\cite{deep_cluster,conv_tasnet} before perception and reasoning.

\subsection{Results on Related Tasks}
\textbf{Results on Ref-AVS.}
As shown in \Cref{tab:refavs}, our \ourinstance achieves SOTA performance on Ref-AVS~\cite{refavs}, greatly surpassing previous methods. Specifically, \ourinstance improves the $\mathcal{J}$ by 17.5\% and 8.8\% on Seen and Unseen splits respectively compared to EEMC~\cite{refavs}, demonstrating good generalization ability. 
While \ourinstance gets suboptimal score on Null split, which may inherit from hallucination in MLLM, EEMC achieves 0.7\% $\mathcal{S}$. We argue that videos in Null split contain empty or fixed audio with unrelated expressions, so EEMC is likely overfiting to identify audio patterns without understanding. Therefore, our model's performance~(9.8\%) is reasonable, and \ourinstance's advantages on other metrics also demonstrate its robustness.

\textbf{Results\!~on~\!Referring~\!Image~\!Segmentation.}
As shown in \Cref{tab:refer_seg_results}, our \ourinstance outperforms previous referring image segmentation methods on refCOCO series \cite{Yu,refcocog}, achieving state-of-the-art performance. 
The performance on ReasonSeg~\cite{lisa} falls short of LISA~\cite{lisa} and VideoLISA~\cite{videolisa}, likely due to the limited reasoning abilities of our smaller 0.5B LLM, which are crucial for this dataset.

\textbf{Results on Referring Video Segmentation.}
As shown in \Cref{tab:rvos_dataset}, the proposed \ourinstance achieves competitive results on MeViS~\cite{mevis} (43.2\%), R-YTVOS~\cite{refer_youtube_vos} (62.1\%), and R-DAVIS$_{17}$~\cite{refer_davis} (65.2\%).~On ReVOS~\cite{visa}, it achieves new state-of-the-art results with 47.3\% $\mathcal{J}\&\mathcal{F}$ and 19.3\% robustness score, surpassing VISA~\cite{visa} by 0.2\% and 4.0\% respectively, demonstrating strong perception and reasoning ability in complex video scenes.

\begin{table}[t]
    \centering
    \footnotesize
    \setlength\tabcolsep{3.96pt}
    \caption{Performance on testing set of \textbf{Ref-AVS}~\cite{refavs}.}
    \vspace{-3mm}
    \label{tab:refavs}
    \begin{tabular}{lccccccc}
        \toprule
        \multirow{2}{*}{\textbf{Method}} & \multicolumn{2}{c}{\textbf{Seen}} & \multicolumn{2}{c}{\textbf{Unseen}} & \multicolumn{2}{c}{\textbf{Mix (S+U)}} & \textbf{Null} \\
        & $\mathcal{J}$ & $\mathcal{F}$ & $\mathcal{J}$ & $\mathcal{F}$ & $\mathcal{J}$ & $\mathcal{F}$ & $\mathcal{S}$ ($\downarrow$) \\
        \myrule
        AVSBench~\cite{avs}          & 23.2 & 51.1 & 32.4 & 54.7 & 27.8 & 52.9 & 20.8 \\
        AVGSegFormer~\cite{avsegformer} & 33.5 & 47.0 & 36.1 & 50.1 & 34.8 & 48.6 & 17.1 \\
        GAVS~\cite{gavs}             & 28.9 & 49.8 & 29.8 & 49.7 & 29.4 & 49.8 & 19.0 \\
        ReferFormer~\cite{referformer} & 31.3 & 50.1 & 30.4 & 48.8 & 30.9 & 49.5 & 17.6 \\
        R2VOS~\cite{r2vos}           & 25.0 & 41.0 & 27.9 & 49.8 & 26.5 & 45.4 & 18.3 \\
        EEMC~\cite{refavs}           & 34.2 & 51.3 & 49.5 & 64.8 & 41.9 & 58.1 & \ \ \textbf{0.7} \\
        \hline
        \rowcolor{cyan!10}
        \textbf{\ourinstance (ours)} & \textbf{51.7} & \textbf{58.7} & \textbf{58.3} & \textbf{65.1} & \textbf{54.5} & \textbf{61.4} & \ \ 9.8 \\
        \myrule
    \end{tabular}
\end{table}

\begin{table}[t]
    \footnotesize
    \centering
    \setlength\tabcolsep{1.7pt}
    \caption{cIoU Results on \textbf{refCOCO}~\cite{Yu,refcocog} and \textbf{ReasonSeg}~\cite{lisa}.}
    \vspace{-3mm}
    \label{tab:refer_seg_results}   
    \resizebox{0.47\textwidth}{!}{
    \begin{tabular}{ l | c c c | c c c | c c | c c }
        \toprule
        \multirow{2}*{\textbf{Method}} & \multicolumn{3}{c}{\textbf{refCOCO}} & \multicolumn{3}{c}{\textbf{refCOCO+}}  & \multicolumn{2}{c}{\textbf{refCOCOg}} & \multicolumn{2}{c}{\textbf{ReasonSeg}}\\ 
        ~ & val & testA & testB & val & testA & testB & val$_{\rm{U}}$ & test$_{\rm{U}}$ & val & test  \\ 
        \myrule
        LAVT~\cite{yang2022lavt} & 72.7 & 75.8 & 68.8 & 62.1 & 68.4 & 55.1 & 61.2 & 62.1 & - & - \\
        ReLA~\cite{gres} & 73.8 & 76.5 & 70.2 & 66.0 & 71.0 & 57.7 & 65.0 & 66.0  & - & -\\
        LISA-7B~\cite{lisa} & 74.1 & 76.5 & 71.1 & 62.4 & 67.4 & 56.5 & 66.4 & 68.5 & 54.0 & 48.4            \\
        LISA-13B~\cite{lisa} & - & - & - & - & - & - & - & - & 62.9 & 51.1            \\
        VISA-7B~\cite{visa} & 72.4 & 75.5 & 68.1 & 59.8 & 64.8 & 53.1 & 65.5 & 66.4 & 57.8 & -   \\
        VideoLISA-3.8B~\cite{videolisa} & 73.8 & 76.6 & 68.8 & 63.4 & 68.8 & 56.2 & 68.3 & 68.8 & \textbf{67.1} & \textbf{54.4}     \\

        \hline
        \rowcolor{cyan!10} 
        \textbf{\ourinstance} (\textbf{ours}) & \textbf{74.2} & \textbf{78.3} & \textbf{72.0} & \textbf{68.8} & \textbf{71.7} & \textbf{62.0} & \textbf{70.6} & \textbf{72.3} & 61.2 & 50.8 \\
        \myrule            
    \end{tabular}
    }
\end{table}

\begin{table}[ht]
    \footnotesize
    \centering
    \caption{Results on \textbf{referring/reasoning video segmentation}~\cite{mevis,refer_youtube_vos,refer_davis,visa}. $\mathcal{J}\&\mathcal{F}$ is the default metric, $\mathcal{R}$ is the robustness score.}
    \setlength\tabcolsep{1.86pt}
    \label{tab:rvos_dataset}
    \vspace{-3mm}
    \resizebox{0.476\textwidth}{!}{
    \begin{tabular}{lccccc}
    \specialrule{.1em}{.05em}{.05em}   
    Method & MeViS & R-YTVOS & R-DAVIS$_{17}$ & ReVOS & ReVOS$_{(\mathcal{R})}$ \\
    \myrule
    ReferFormer~\cite{referformer}  & 31.0 & 62.9 & 61.1 & 14.9     &\ \ 4.9     \\
    MTTR~\cite{MTTR}      & 30.0 & 55.3 &  -        &  25.5   &\ \   5.6  \\
    LMPM~\cite{mevis}        & 37.2 &    - &    - &  26.4    &\ \   3.2    \\
    \hline
    LISA-7B~\cite{lisa}     & - & 50.2 & 58.4 &  40.9    &\ \  9.3     \\
    TrackGPT-7B~\cite{trackgpt} & 40.1 & 56.4 & 63.2 &  43.6    & 11.6     \\
    VideoLISA-3.8B~\cite{videolisa}   & 42.3 & 61.7 & 67.7 & - & -\\
    VISA-7B~\cite{visa}     & \textbf{44.5} & \textbf{63.0} & \textbf{70.4} &  47.1    & 15.3     \\
    \hline
    \rowcolor{cyan!10} 
    \textbf{\ourinstance} (\textbf{ours})     &  43.2   & 62.1   &   65.2     & \textbf{47.3}     & \textbf{19.3}     \\
    \myrule 
    \end{tabular}%
    }
    \vspace{-2mm}
\end{table}

\section{Conclusion and Discussion}
\label{sec:conclusion}
We introduce \ourdataset, a large-scale dataset supporting advanced multimodal referring and reasoning in audiovisual scenes with 8 types of expressions across text, speech, sound, and image. Compared to previous referring datasets focused on salient video or sound occurrence, \ourdataset enables deeper understanding and reasoning of audiovisual content. 
To address the challenges posed by \ourdataset, we propose a simple yet effective baseline \ourmodel, a segmentation model based on multimodal large language models.
Our \ourmodel leverages audio-visual interleaving for temporal alignment and query propagation for efficient segmentation, enabling effective handling of multimodal inputs without additional parameters.
~Extensive experiments on 10 datasets demonstrate the effectiveness of our \ourmodel.

\textbf{Future Directions.}
Although our baseline \ourmodel has shown promising results on \ourdataset, there is still much room for improvement:
\textbf{i)} Exploring more efficient audio-visual fusion methods, such as learning more robust audio-visual representations.
\textbf{ii)} Disentangling sounds when multiple objects make sounds simultaneously, which is crucial for subsequent reasoning.
\textbf{iii)} How to effectively combine multiple modalities in expressions by using joint representations or cross-modal fusion.
\textbf{iv)} Exploring more advanced segmentation models to handle complex scenarios, \eg, occlusion, disappearance and reappearance.
\textbf{v)} Pre-training \ourmodel on larger audio-visual datasets to further improve its generalization to unseen tasks.
\textbf{vi)} Developing more interactive and conversational capabilities to enable multi-round interactions with users, allowing for clarification and refinement of referring expressions.

\footnotesize{\paragraph{Acknowledgement.}~This project was supported by the National Natural Science Foundation of China (NSFC) under Grant No. 62472104.}

{
    \small
    \bibliographystyle{ieeenat_fullname}
    \bibliography{main}

\begin{thebibliography}{87}
\providecommand{\natexlab}[1]{#1}
\providecommand{\url}[1]{\texttt{#1}}
\expandafter\ifx\csname urlstyle\endcsname\relax
  \providecommand{\doi}[1]{doi: #1}\else
  \providecommand{\doi}{doi: \begingroup \urlstyle{rm}\Url}\fi

\bibitem[Bai et~al.(2023{\natexlab{a}})Bai, Bai, Chu, Cui, Dang, Deng, Fan, Ge, Han, Huang, et~al.]{qwen}
Jinze Bai, Shuai Bai, Yunfei Chu, Zeyu Cui, Kai Dang, Xiaodong Deng, Yang Fan, Wenbin Ge, Yu Han, Fei Huang, et~al.
\newblock {Qwen Technical Report}.
\newblock \emph{arXiv}, 2023{\natexlab{a}}.

\bibitem[Bai et~al.(2023{\natexlab{b}})Bai, Bai, Yang, Wang, Tan, Wang, Lin, Zhou, and Zhou]{qwenvl}
Jinze Bai, Shuai Bai, Shusheng Yang, Shijie Wang, Sinan Tan, Peng Wang, Junyang Lin, Chang Zhou, and Jingren Zhou.
\newblock {Qwen-VL: A Frontier Large Vision-Language Model with Versatile Abilities}.
\newblock \emph{arXiv preprint arXiv:2308.12966}, 2023{\natexlab{b}}.

\bibitem[Bai et~al.(2024)Bai, He, Mei, WANG, Gao, Chen, liulei, Zhang, and Shou]{videolisa}
Zechen Bai, Tong He, Haiyang Mei, Pichao WANG, Ziteng Gao, Joya Chen, liulei, Zheng Zhang, and Mike~Zheng Shou.
\newblock {One Token to Seg Them All: Language Instructed Reasoning Segmentation in Videos}.
\newblock In \emph{Adv. Neural Inform. Process. Syst.}, 2024.

\bibitem[Banerjee and Lavie(2005)]{meteor}
Satanjeev Banerjee and Alon Lavie.
\newblock {METEOR: An Automatic Metric for MT Evaluation with Improved Correlation with Human Judgments}.
\newblock In \emph{Assoc. Comput. Linguist. Worksh.}, 2005.

\bibitem[Botach et~al.(2022)Botach, Zheltonozhskii, and Baskin]{MTTR}
Adam Botach, Evgenii Zheltonozhskii, and Chaim Baskin.
\newblock {End-to-End Referring Video Object Segmentation with Multimodal Transformers}.
\newblock In \emph{IEEE Conf. Comput. Vis. Pattern Recog.}, 2022.

\bibitem[Bregman(1994)]{bregman1994auditory}
Albert~S Bregman.
\newblock \emph{{Auditory Scene Analysis: The Perceptual Organization of Sound}}.
\newblock MIT press, 1994.

\bibitem[Caesar et~al.(2018)Caesar, Uijlings, and Ferrari]{cocostuff}
Holger Caesar, Jasper Uijlings, and Vittorio Ferrari.
\newblock {COCO-Stuff: Thing and Stuff Classes in Context}.
\newblock In \emph{IEEE Conf. Comput. Vis. Pattern Recog.}, 2018.

\bibitem[Chalk et~al.(2024)Chalk, Huh, Kazakos, Zisserman, and Damen]{actionrecognition}
Jacob Chalk, Jaesung Huh, Evangelos Kazakos, Andrew Zisserman, and Dima Damen.
\newblock {TIM: A Time Interval Machine for Audio-Visual Action Recognition}.
\newblock In \emph{IEEE Conf. Comput. Vis. Pattern Recog.}, 2024.

\bibitem[Chen et~al.(2021)Chen, Chai, et~al.]{gigaspeech}
Guoguo Chen, Shuzhou Chai, et~al.
\newblock {GigaSpeech: An Evolving, Multi-domain ASR Corpus with 10,000 Hours of Transcribed Audio}.
\newblock In \emph{Proc. Interspeech 2021}, 2021.

\bibitem[Chen et~al.(2020)Chen, Xie, Vedaldi, and Zisserman]{vggsound}
Honglie Chen, Weidi Xie, Andrea Vedaldi, and Andrew Zisserman.
\newblock {VGGSound: A Large-scale Audio-Visual Dataset}.
\newblock In \emph{IEEE Int. Conf. Acoust. Speech Signal Process.}, 2020.

\bibitem[Chen et~al.(2014)Chen, Mottaghi, Liu, Fidler, Urtasun, and Yuille]{pascalpart}
Xianjie Chen, Roozbeh Mottaghi, Xiaobai Liu, Sanja Fidler, Raquel Urtasun, and Alan Yuille.
\newblock {Detect What You Can: Detecting and Representing Objects using Holistic Models and Body Parts}.
\newblock In \emph{IEEE Conf. Comput. Vis. Pattern Recog.}, 2014.

\bibitem[Chen et~al.(2023)Chen, Duan, Wang, He, Lu, Dai, and Qiao]{vitadapter}
Zhe Chen, Yuchen Duan, Wenhai Wang, Junjun He, Tong Lu, Jifeng Dai, and Yu Qiao.
\newblock {Vision Transformer Adapter for Dense Predictions}.
\newblock In \emph{Int. Conf. Learn. Represent.}, 2023.

\bibitem[Chen et~al.(2024)Chen, Wang, Tian, Ye, Gao, Cui, Tong, Hu, Luo, Ma, et~al.]{internvl}
Zhe Chen, Weiyun Wang, Hao Tian, Shenglong Ye, Zhangwei Gao, Erfei Cui, Wenwen Tong, Kongzhi Hu, Jiapeng Luo, Zheng Ma, et~al.
\newblock {How Far Are We to GPT-4V? Closing the Gap to Commercial Multimodal Models with Open-Source Suites}.
\newblock \emph{arXiv preprint arXiv:2404.16821}, 2024.

\bibitem[Cheng et~al.(2022)Cheng, Misra, Schwing, Kirillov, and Girdhar]{mask2former}
Bowen Cheng, Ishan Misra, Alexander~G Schwing, Alexander Kirillov, and Rohit Girdhar.
\newblock {Masked-attention Mask Transformer for Universal Image Segmentation}.
\newblock In \emph{IEEE Conf. Comput. Vis. Pattern Recog.}, 2022.

\bibitem[Cheng et~al.(2024)Cheng, Leng, Zhang, Xin, Li, Chen, Zhu, Zhang, Luo, Zhao, et~al.]{videollama2}
Zesen Cheng, Sicong Leng, Hang Zhang, Yifei Xin, Xin Li, Guanzheng Chen, Yongxin Zhu, Wenqi Zhang, Ziyang Luo, Deli Zhao, et~al.
\newblock {VideoLLaMA 2: Advancing Spatial-Temporal Modeling and Audio Understanding in Video-LLMs}.
\newblock \emph{arXiv preprint arXiv:2406.07476}, 2024.

\bibitem[Chu et~al.(2024)Chu, Xu, Yang, Wei, Wei, Guo, Leng, Lv, He, Lin, et~al.]{qwen2audio}
Yunfei Chu, Jin Xu, Qian Yang, Haojie Wei, Xipin Wei, Zhifang Guo, Yichong Leng, Yuanjun Lv, Jinzheng He, Junyang Lin, et~al.
\newblock {Qwen2-Audio Technical Report}.
\newblock \emph{arXiv preprint arXiv:2407.10759}, 2024.

\bibitem[Ding et~al.(2023{\natexlab{a}})Ding, Liu, He, Jiang, and Loy]{mevis}
Henghui Ding, Chang Liu, Shuting He, Xudong Jiang, and Chen~Change Loy.
\newblock {MeViS: A Large-scale Benchmark for Video Segmentation with Motion Expressions}.
\newblock In \emph{Int. Conf. Comput. Vis.}, 2023{\natexlab{a}}.

\bibitem[Ding et~al.(2023{\natexlab{b}})Ding, Liu, He, Jiang, Torr, and Bai]{mose}
Henghui Ding, Chang Liu, Shuting He, Xudong Jiang, Philip~HS Torr, and Song Bai.
\newblock {MOSE: A New Dataset for Video Object Segmentation in Complex Scenes}.
\newblock In \emph{Int. Conf. Comput. Vis.}, 2023{\natexlab{b}}.

\bibitem[Ding et~al.(2025{\natexlab{a}})Ding, Tang, He, Liu, Wu, and Jiang]{ReferringSurvey}
Henghui Ding, Song Tang, Shuting He, Chang Liu, Zuxuan Wu, and Yu-Gang Jiang.
\newblock Multimodal referring segmentation: A survey.
\newblock \emph{arXiv}, 2025{\natexlab{a}}.

\bibitem[Ding et~al.(2025{\natexlab{b}})Ding, Ying, Liu, He, Jiang, Torr, and Bai]{MOSEv2}
Henghui Ding, Kaining Ying, Chang Liu, Shuting He, Yu-Gang Jiang, Philip~HS Torr, and Song Bai.
\newblock {MOSEv2}: A more challenging dataset for video object segmentation in complex scenes.
\newblock \emph{arXiv}, 2025{\natexlab{b}}.

\bibitem[Fu et~al.(2024)Fu, Lin, Long, Shen, Zhao, Zhang, Wang, Yin, Ma, Zheng, et~al.]{vita}
Chaoyou Fu, Haojia Lin, Zuwei Long, Yunhang Shen, Meng Zhao, Yifan Zhang, Xiong Wang, Di Yin, Long Ma, Xiawu Zheng, et~al.
\newblock {VITA: Towards Open-Source Interactive Omni Multimodal LLM}.
\newblock \emph{arXiv}, 2024.

\bibitem[Gao et~al.(2024)Gao, Chen, Chen, Wang, and Lu]{avsegformer}
Shengyi Gao, Zhe Chen, Guo Chen, Wenhai Wang, and Tong Lu.
\newblock {AVSegFormer: Audio-Visual Segmentation with Transformer}.
\newblock In \emph{AAAI}, 2024.

\bibitem[GPT-SoVITS(2024)]{gpt_sovits}
GPT-SoVITS.
\newblock \url{https://github.com/RVC-Boss/GPT-SoVITS}, 2024.

\bibitem[Guo et~al.(2024)Guo, Qu, Niu, Qi, Yue, Shi, Xing, and Ying]{ovavss}
Ruohao Guo, Liao Qu, Dantong Niu, Yanyu Qi, Wenzhen Yue, Ji Shi, Bowei Xing, and Xianghua Ying.
\newblock {Open-Vocabulary Audio-Visual Semantic Segmentation}.
\newblock In \emph{ACM Int. Conf. Multimedia}, 2024.

\bibitem[He et~al.(2024)He, Wang, Wang, Lu, He, Lan, Luo, and Xie]{anyref}
Junwen He, Yifan Wang, Lijun Wang, Huchuan Lu, Jun-Yan He, Jin-Peng Lan, Bin Luo, and Xuansong Xie.
\newblock {Multi-modal Instruction Tuned LLMs with Fine-grained Visual Perception}.
\newblock In \emph{IEEE Conf. Comput. Vis. Pattern Recog.}, 2024.

\bibitem[He and Ding(2024)]{dshmp}
Shuting He and Henghui Ding.
\newblock {Decoupling Static and Hierarchical Motion Perception for Referring Video Segmentation}.
\newblock In \emph{IEEE Conf. Comput. Vis. Pattern Recog.}, 2024.

\bibitem[Heo et~al.(2023)Heo, Hwang, Hyun, Kim, Oh, Lee, and Kim]{genvis}
Miran Heo, Sukjun Hwang, Jeongseok Hyun, Hanjung Kim, Seoung~Wug Oh, Joon-Young Lee, and Seon~Joo Kim.
\newblock {A Generalized Framework for Video Instance Segmentation}.
\newblock In \emph{IEEE Conf. Comput. Vis. Pattern Recog.}, 2023.

\bibitem[Hershey et~al.(2016)Hershey, Chen, Le~Roux, and Watanabe]{deep_cluster}
John~R Hershey, Zhuo Chen, Jonathan Le~Roux, and Shinji Watanabe.
\newblock {Deep clustering: Discriminative embeddings for segmentation and separation}.
\newblock In \emph{IEEE Int. Conf. Acoust. Speech Signal Process.}, 2016.

\bibitem[Hu et~al.(2022)Hu, Shen, Wallis, Allen-Zhu, Li, Wang, Wang, and Chen]{lora}
Edward~J Hu, Yelong Shen, Phillip Wallis, Zeyuan Allen-Zhu, Yuanzhi Li, Shean Wang, Lu Wang, and Weizhu Chen.
\newblock {LoRA: Low-Rank Adaptation of Large Language Models}.
\newblock In \emph{Int. Conf. Learn. Represent.}, 2022.

\bibitem[Huang et~al.(2023)Huang, Tian, Kumar, and Xu]{objectlocalization}
Chao Huang, Yapeng Tian, Anurag Kumar, and Chenliang Xu.
\newblock {Egocentric Audio-Visual Object Localization}.
\newblock In \emph{IEEE Conf. Comput. Vis. Pattern Recog.}, 2023.

\bibitem[Khoreva et~al.(2019)Khoreva, Rohrbach, and Schiele]{refer_davis}
Anna Khoreva, Anna Rohrbach, and Bernt Schiele.
\newblock {Video Object Segmentation with Language Referring Expressions}.
\newblock In \emph{ACCV}, 2019.

\bibitem[Kirillov et~al.(2023)Kirillov, Mintun, Ravi, Mao, Rolland, Gustafson, Xiao, Whitehead, Berg, Lo, et~al.]{sam}
Alexander Kirillov, Eric Mintun, Nikhila Ravi, Hanzi Mao, Chloe Rolland, Laura Gustafson, Tete Xiao, Spencer Whitehead, Alexander~C Berg, Wan-Yen Lo, et~al.
\newblock {Segment Anything}.
\newblock In \emph{Int. Conf. Comput. Vis.}, 2023.

\bibitem[Lai et~al.(2024)Lai, Tian, Chen, Li, Yuan, Liu, and Jia]{lisa}
Xin Lai, Zhuotao Tian, Yukang Chen, Yanwei Li, Yuhui Yuan, Shu Liu, and Jiaya Jia.
\newblock {LISA: Reasoning Segmentation via Large Language Model}.
\newblock In \emph{IEEE Conf. Comput. Vis. Pattern Recog.}, 2024.

\bibitem[Lei et~al.(2018)Lei, Yu, Bansal, and Berg]{tvqa}
Jie Lei, Licheng Yu, Mohit Bansal, and Tamara~L Berg.
\newblock {TVQA: Localized, Compositional Video Question Answering}.
\newblock In \emph{Proc. of the Conf. on Empirical Methods in Nat. Lang. Process.}, 2018.

\bibitem[Li et~al.(2022)Li, Wei, Tian, Xu, Wen, and Hu]{music_avqa}
Guangyao Li, Yake Wei, Yapeng Tian, Chenliang Xu, Ji-Rong Wen, and Di Hu.
\newblock {Learning to Answer Questions in Dynamic Audio-Visual Scenarios}.
\newblock In \emph{IEEE Conf. Comput. Vis. Pattern Recog.}, 2022.

\bibitem[Li et~al.(2024{\natexlab{a}})Li, Du, and Hu]{hu_4}
Guangyao Li, Henghui Du, and Di Hu.
\newblock {Boosting Audio Visual Question Answering via Key Semantic-Aware Cues}.
\newblock In \emph{ACM Int. Conf. Multimedia}, pages 5997--6005, 2024{\natexlab{a}}.

\bibitem[Li et~al.(2025)Li, Zhao, Huang, Guo, and Tian]{hu_5}
Jia Li, Wenjie Zhao, Ziru Huang, Yunhui Guo, and Yapeng Tian.
\newblock {Do Audio-Visual Segmentation Models Truly Segment Sounding Objects?}
\newblock \emph{arXiv}, 2025.

\bibitem[Li et~al.(2023)Li, Wang, Xu, Li, Raj, and Lu]{r2vos}
Xiang Li, Jinglu Wang, Xiaohao Xu, Xiao Li, Bhiksha Raj, and Yan Lu.
\newblock {Robust Referring Video Object Segmentation with Cyclic Structural Consensus}.
\newblock In \emph{Int. Conf. Comput. Vis.}, 2023.

\bibitem[Li et~al.(2024{\natexlab{b}})Li, Sun, Lin, Li, Dong, Zhang, Ding, Song, Cheng, Huo, et~al.]{baichuan_omni}
Yadong Li, Haoze Sun, Mingan Lin, Tianpeng Li, Guosheng Dong, Tao Zhang, Bowen Ding, Wei Song, Zhenglin Cheng, Yuqi Huo, et~al.
\newblock {Baichuan-Omni Technical Report}.
\newblock \emph{arXiv preprint arXiv:2410.08565}, 2024{\natexlab{b}}.

\bibitem[{Linfeng Yuan and Miaojing Shi and Zijie Yue and Qijun Chen}(2024)]{losh}
{Linfeng Yuan and Miaojing Shi and Zijie Yue and Qijun Chen}.
\newblock Losh: Long-short text joint prediction network for referring video object segmentation.
\newblock In \emph{IEEE Conf. Comput. Vis. Pattern Recog.}, 2024.

\bibitem[Liu et~al.(2023{\natexlab{a}})Liu, Ding, and Jiang]{gres}
Chang Liu, Henghui Ding, and Xudong Jiang.
\newblock {GRES: Generalized Referring Expression Segmentation}.
\newblock In \emph{IEEE Conf. Comput. Vis. Pattern Recog.}, 2023{\natexlab{a}}.

\bibitem[Liu et~al.(2024{\natexlab{a}})Liu, Jiang, and Ding]{primitivenet}
Chang Liu, Xudong Jiang, and Henghui Ding.
\newblock Primitivenet: decomposing the global constraints for referring segmentation.
\newblock \emph{Visual Intelligence}, 2024{\natexlab{a}}.

\bibitem[Liu et~al.(2023{\natexlab{b}})Liu, Li, Wu, and Lee]{llava}
Haotian Liu, Chunyuan Li, Qingyang Wu, and Yong~Jae Lee.
\newblock {Visual Instruction Tuning}.
\newblock In \emph{Adv. Neural Inform. Process. Syst.}, 2023{\natexlab{b}}.

\bibitem[Liu et~al.(2024{\natexlab{b}})Liu, Li, Li, and Lee]{llavav1.5}
Haotian Liu, Chunyuan Li, Yuheng Li, and Yong~Jae Lee.
\newblock {Improved Baselines with Visual Instruction Tuning}.
\newblock In \emph{IEEE Conf. Comput. Vis. Pattern Recog.}, 2024{\natexlab{b}}.

\bibitem[Liu et~al.(2024{\natexlab{c}})Liu, Ying, Zhang, Yang, Lin, Zhang, Li, Qiao, Luo, Shao, et~al.]{convbench}
Shuo Liu, Kaining Ying, Hao Zhang, Yue Yang, Yuqi Lin, Tianle Zhang, Chuanhao Li, Yu Qiao, Ping Luo, Wenqi Shao, et~al.
\newblock {ConvBench: A Multi-Turn Conversation Evaluation Benchmark with Hierarchical Ablation Capability for Large Vision-Language Models}.
\newblock In \emph{Adv. Neural Inform. Process. Syst. D\&B}, 2024{\natexlab{c}}.

\bibitem[Luo and Mesgarani(2019)]{conv_tasnet}
Yi Luo and Nima Mesgarani.
\newblock {Conv-TasNet: Surpassing Ideal Time-Frequency Magnitude Masking for Speech Separation}.
\newblock \emph{IEEE/ACM Trans. Audio Speech Lang. Process.}, 27\penalty0 (8), 2019.

\bibitem[Ma et~al.(2024)Ma, Sun, Wang, and Hu]{hu_3}
Juncheng Ma, Peiwen Sun, Yaoting Wang, and Di Hu.
\newblock {Stepping Stones: A Progressive Training Strategy for Audio-Visual Semantic Segmentation}.
\newblock In \emph{Eur. Conf. Comput. Vis.}, 2024.

\bibitem[Mao et~al.(2016)Mao, Huang, Toshev, Camburu, Yuille, and Murphy]{refcocog}
Junhua Mao, Jonathan Huang, Alexander Toshev, Oana Camburu, Alan~L Yuille, and Kevin Murphy.
\newblock {Generation and Comprehension of Unambiguous Object Descriptions}.
\newblock In \emph{IEEE Conf. Comput. Vis. Pattern Recog.}, 2016.

\bibitem[Milletari et~al.(2016)Milletari, Navab, and Ahmadi]{diceloss}
Fausto Milletari, Nassir Navab, and Seyed-Ahmad Ahmadi.
\newblock {V-Net: Fully Convolutional Neural Networks for Volumetric Medical Image Segmentation}.
\newblock In \emph{IEEE Int. Conf. 3D Vis.}, 2016.

\bibitem[OpenAI(2023)]{openai_tts}
OpenAI.
\newblock \url{https://platform.openai.com/docs/guides/text-to-speech}, 2023.

\bibitem[OpenAI(2024)]{gpt4o}
OpenAI.
\newblock \url{https://openai.com/index/hello-gpt-4o}, 2024.

\bibitem[Pan et~al.(2022)Pan, Shi, Zhao, Zhu, He, Pan, Gao, Yu, Wu, and Tian]{wnet}
Wenwen Pan, Haonan Shi, Zhou Zhao, Jieming Zhu, Xiuqiang He, Zhigeng Pan, Lianli Gao, Jun Yu, Fei Wu, and Qi Tian.
\newblock {Wnet: Audio-Guided Video Object Segmentation via Wavelet-Based Cross-Modal Denoising Networks}.
\newblock In \emph{IEEE Conf. Comput. Vis. Pattern Recog.}, 2022.

\bibitem[Pi et~al.(2023)Pi, Gao, Diao, Pan, Dong, Zhang, Yao, Han, Xu, Kong, et~al.]{detgpt}
Renjie Pi, Jiahui Gao, Shizhe Diao, Rui Pan, Hanze Dong, Jipeng Zhang, Lewei Yao, Jianhua Han, Hang Xu, Lingpeng Kong, et~al.
\newblock {DetGPT: Detect What You Need via Reasoning}.
\newblock In \emph{Proc. of the Conf. on Empirical Methods in Nat. Lang. Process.}, 2023.

\bibitem[Radford et~al.(2023)Radford, Kim, Xu, Brockman, McLeavey, and Sutskever]{whisper}
Alec Radford, Jong~Wook Kim, Tao Xu, Greg Brockman, Christine McLeavey, and Ilya Sutskever.
\newblock {Robust Speech Recognition via Large-Scale Weak Supervision}.
\newblock In \emph{Int. Conf. Mach. Learn.}, 2023.

\bibitem[Ramanathan et~al.(2023)Ramanathan, Kalia, Petrovic, Wen, Zheng, Guo, Wang, Marquez, Kovvuri, Kadian, et~al.]{pacolvis}
Vignesh Ramanathan, Anmol Kalia, Vladan Petrovic, Yi Wen, Baixue Zheng, Baishan Guo, Rui Wang, Aaron Marquez, Rama Kovvuri, Abhishek Kadian, et~al.
\newblock {PACO: Parts and Attributes of Common Objects}.
\newblock In \emph{IEEE Conf. Comput. Vis. Pattern Recog.}, 2023.

\bibitem[Ravi et~al.(2024)Ravi, Gabeur, Hu, Hu, Ryali, Ma, Khedr, R{\"a}dle, Rolland, Gustafson, et~al.]{sam2}
Nikhila Ravi, Valentin Gabeur, Yuan-Ting Hu, Ronghang Hu, Chaitanya Ryali, Tengyu Ma, Haitham Khedr, Roman R{\"a}dle, Chloe Rolland, Laura Gustafson, et~al.
\newblock {SAM 2: Segment Anything in Images and Videos}.
\newblock \emph{arXiv preprint arXiv:2408.00714}, 2024.

\bibitem[Seo et~al.(2020)Seo, Lee, and Han]{refer_youtube_vos}
Seonguk Seo, Joon-Young Lee, and Bohyung Han.
\newblock {URVOS: Unified Referring Video Object Segmentation Network with a Large-Scale Benchmark}.
\newblock In \emph{Eur. Conf. Comput. Vis.}, 2020.

\bibitem[Sun et~al.(2024{\natexlab{a}})Sun, Yu, Tang, Chen, Tan, Li, Lu, Ma, Wang, and Zhang]{videosalmonn}
Guangzhi Sun, Wenyi Yu, Changli Tang, Xianzhao Chen, Tian Tan, Wei Li, Lu Lu, Zejun Ma, Yuxuan Wang, and Chao Zhang.
\newblock {video-SALMONN: Speech-Enhanced Audio-Visual Large Language Models}.
\newblock In \emph{Int. Conf. Mach. Learn.}, 2024{\natexlab{a}}.

\bibitem[Sun et~al.(2024{\natexlab{b}})Sun, Xu, Wu, and Xie]{autoacd}
Luoyi Sun, Xuenan Xu, Mengyue Wu, and Weidi Xie.
\newblock {Auto-ACD: A Large-scale Dataset for Audio-Language Representation Learning}.
\newblock In \emph{ACM Int. Conf. Multimedia}, 2024{\natexlab{b}}.

\bibitem[Sun et~al.(2024{\natexlab{c}})Sun, Zhang, and Hu]{hu_6}
Peiwen Sun, Honggang Zhang, and Di Hu.
\newblock {Unveiling and Mitigating Bias in Audio Visual Segmentation}.
\newblock In \emph{ACM Int. Conf. Multimedia}, 2024{\natexlab{c}}.

\bibitem[Tang et~al.(2023)Tang, Yu, Sun, Chen, Tan, Li, Lu, Ma, and Zhang]{salmonn}
Changli Tang, Wenyi Yu, Guangzhi Sun, Xianzhao Chen, Tian Tan, Wei Li, Lu Lu, Zejun Ma, and Chao Zhang.
\newblock {SALMONN: Towards Generic Hearing Abilities for Large Language Models}.
\newblock In \emph{Int. Conf. Learn. Represent.}, 2023.

\bibitem[Tian et~al.(2018)Tian, Shi, Li, Duan, and Xu]{eventlocalization}
Yapeng Tian, Jing Shi, Bochen Li, Zhiyao Duan, and Chenliang Xu.
\newblock {Audio-Visual Event Localization in Unconstrained Videos}.
\newblock In \emph{Eur. Conf. Comput. Vis.}, 2018.

\bibitem[Wang et~al.(2024{\natexlab{a}})Wang, Bai, Tan, Wang, Fan, Bai, Chen, Liu, Wang, Ge, et~al.]{qwen2vl}
Peng Wang, Shuai Bai, Sinan Tan, Shijie Wang, Zhihao Fan, Jinze Bai, Keqin Chen, Xuejing Liu, Jialin Wang, Wenbin Ge, et~al.
\newblock {Qwen2-VL: Enhancing Vision-Language Model's Perception of the World at Any Resolution}.
\newblock \emph{arXiv preprint arXiv:2409.12191}, 2024{\natexlab{a}}.

\bibitem[Wang et~al.(2022)Wang, Wang, Wang, Guo, Liu, and Sun]{wang2022audio}
Yefei Wang, Kaili Wang, Yi Wang, Di Guo, Huaping Liu, and Fuchun Sun.
\newblock {Audio-Visual Grounding Referring Expression for Robotic Manipulation}.
\newblock In \emph{IEEE Int. Conf. Robot. Autom.}, 2022.

\bibitem[Wang et~al.(2024{\natexlab{b}})Wang, Liu, Li, Ding, Hu, and Li]{gavs}
Yaoting Wang, Weisong Liu, Guangyao Li, Jian Ding, Di Hu, and Xi Li.
\newblock {Prompting Segmentation with Sound Is Generalizable Audio-Visual Source Localizer}.
\newblock In \emph{AAAI}, 2024{\natexlab{b}}.

\bibitem[Wang et~al.(2024{\natexlab{c}})Wang, Liu, Li, Ding, Hu, and Li]{hu_2}
Yaoting Wang, Weisong Liu, Guangyao Li, Jian Ding, Di Hu, and Xi Li.
\newblock {Prompting Segmentation with Sound Is Generalizable Audio-Visual Source Localizer}.
\newblock In \emph{AAAI}, 2024{\natexlab{c}}.

\bibitem[Wang et~al.(2024{\natexlab{d}})Wang, Sun, Li, Zhang, and Hu]{hu_1}
Yaoting Wang, Peiwen Sun, Yuanchao Li, Honggang Zhang, and Di Hu.
\newblock {Can Textual Semantics Mitigate Sounding Object Segmentation Preference?}
\newblock In \emph{Eur. Conf. Comput. Vis.}, 2024{\natexlab{d}}.

\bibitem[Wang et~al.(2024{\natexlab{e}})Wang, Sun, Zhou, Li, Zhang, and Hu]{refavs}
Yaoting Wang, Peiwen Sun, Dongzhan Zhou, Guangyao Li, Honggang Zhang, and Di Hu.
\newblock {Ref-AVS: Refer and Segment Objects in Audio-Visual Scenes}.
\newblock In \emph{Eur. Conf. Comput. Vis.}, 2024{\natexlab{e}}.

\bibitem[Wu et~al.(2022)Wu, Jiang, Sun, Yuan, and Luo]{referformer}
Jiannan Wu, Yi Jiang, Peize Sun, Zehuan Yuan, and Ping Luo.
\newblock {Language as Queries for Referring Video Object Segmentation}.
\newblock In \emph{IEEE Conf. Comput. Vis. Pattern Recog.}, 2022.

\bibitem[Yan et~al.(2024{\natexlab{a}})Yan, Wang, Yan, Jiang, Hu, Kang, Xie, and Gavves]{visa}
Cilin Yan, Haochen Wang, Shilin Yan, Xiaolong Jiang, Yao Hu, Guoliang Kang, Weidi Xie, and Efstratios Gavves.
\newblock {VISA: Reasoning Video Object Segmentation via Large Language Models}.
\newblock In \emph{Eur. Conf. Comput. Vis.}, 2024{\natexlab{a}}.

\bibitem[Yan et~al.(2024{\natexlab{b}})Yan, Zhang, Guo, Chen, Zhang, Li, Qiao, Dong, He, and Gao]{mutr}
Shilin Yan, Renrui Zhang, Ziyu Guo, Wenchao Chen, Wei Zhang, Hongyang Li, Yu Qiao, Hao Dong, Zhongjiang He, and Peng Gao.
\newblock {Referred by Multi-Modality: A Unified Temporal Transformer for Video Object Segmentation}.
\newblock In \emph{AAAI}, 2024{\natexlab{b}}.

\bibitem[Yang et~al.(2022{\natexlab{a}})Yang, Wang, Duan, Chen, Hou, Jin, and Zhu]{avqa}
Pinci Yang, Xin Wang, Xuguang Duan, Hong Chen, Runze Hou, Cong Jin, and Wenwu Zhu.
\newblock {AVQA: A Dataset for Audio-Visual Question Answering on Videos}.
\newblock In \emph{ACM Int. Conf. Multimedia}, 2022{\natexlab{a}}.

\bibitem[Yang et~al.(2022{\natexlab{b}})Yang, Wang, Tang, Chen, Zhao, and Torr]{yang2022lavt}
Zhao Yang, Jiaqi Wang, Yansong Tang, Kai Chen, Hengshuang Zhao, and Philip~HS Torr.
\newblock {LAVT: Language-Aware Vision Transformer for Referring Image Segmentation}.
\newblock In \emph{IEEE Conf. Comput. Vis. Pattern Recog.}, 2022{\natexlab{b}}.

\bibitem[Ying et~al.(2022)Ying, Wang, Bai, and Zhou]{isda}
Kaining Ying, Zhenhua Wang, Cong Bai, and Pengfei Zhou.
\newblock Isda: Position-aware instance segmentation with deformable attention.
\newblock In \emph{IEEE Int. Conf. Acoust. Speech Signal Process.}, 2022.

\bibitem[Ying et~al.(2023)Ying, Zhong, Mao, Wang, Chen, Wu, Liu, Fan, Zhuge, and Shen]{ctvis}
Kaining Ying, Qing Zhong, Weian Mao, Zhenhua Wang, Hao Chen, Lin~Yuanbo Wu, Yifan Liu, Chengxiang Fan, Yunzhi Zhuge, and Chunhua Shen.
\newblock {CTVIS: Consistent Training for Online Video Instance Segmentation}.
\newblock In \emph{Int. Conf. Comput. Vis.}, 2023.

\bibitem[Ying et~al.(2024)Ying, Meng, Wang, Li, Lin, Yang, Zhang, Zhang, Lin, Liu, Lei, Lu, Chen, Xu, Zhang, Zhang, Gao, Wang, Qiao, Luo, Zhang, and Shao]{mmtbench}
Kaining Ying, Fanqing Meng, Jin Wang, Zhiqian Li, Han Lin, Yue Yang, Hao Zhang, Wenbo Zhang, Yuqi Lin, Shuo Liu, Jiayi Lei, Quanfeng Lu, Runjian Chen, Peng Xu, Renrui Zhang, Haozhe Zhang, Peng Gao, Yali Wang, Yu Qiao, Ping Luo, Kaipeng Zhang, and Wenqi Shao.
\newblock {MMT-Bench: A Comprehensive Multimodal Benchmark for Evaluating Large Vision-Language Models Towards Multitask AGI}.
\newblock In \emph{Int. Conf. Mach. Learn.}, 2024.

\bibitem[Ying et~al.(2025)Ying, Hu, and Ding]{MOVE}
Kaining Ying, Hengrui Hu, and Henghui Ding.
\newblock {MOVE}: Motion-guided few-shot video object segmentation.
\newblock In \emph{Int. Conf. Comput. Vis.}, 2025.

\bibitem[Yu et~al.(2016)Yu, Poirson, Yang, Berg, and Berg]{Yu}
Licheng Yu, Patrick Poirson, Shan Yang, Alexander~C Berg, and Tamara~L Berg.
\newblock {Modeling Context in Referring Expressions}.
\newblock In \emph{Eur. Conf. Comput. Vis.}, 2016.

\bibitem[Zeng et~al.(2022)Zeng, Dong, Zhang, Wang, Zhang, and Wei]{motr}
Fangao Zeng, Bin Dong, Yuang Zhang, Tiancai Wang, Xiangyu Zhang, and Yichen Wei.
\newblock {MOTR: End-to-End Multiple-Object Tracking with Transformer}.
\newblock In \emph{Eur. Conf. Comput. Vis.}, 2022.

\bibitem[Zhang et~al.(2023{\natexlab{a}})Zhang, Li, and Bing]{videollama}
Hang Zhang, Xin Li, and Lidong Bing.
\newblock {Video-LLaMA: An Instruction-tuned Audio-Visual Language Model for Video Understanding}.
\newblock In \emph{Proc. of the Conf. on Empirical Methods in Nat. Lang. Process.}, 2023{\natexlab{a}}.

\bibitem[Zhang et~al.(2024)Zhang, Han, Liu, Zhou, Lu, Qiao, Li, and Gao]{llama-adapter}
Renrui Zhang, Jiaming Han, Chris Liu, Aojun Zhou, Pan Lu, Yu Qiao, Hongsheng Li, and Peng Gao.
\newblock {LLaMA-Adapter: Efficient Fine-tuning of Large Language Models with Zero-initialized Attention}.
\newblock In \emph{Int. Conf. Learn. Represent.}, 2024.

\bibitem[Zhang et~al.(2023{\natexlab{b}})Zhang, Sun, Chen, Xiao, Shao, Zhang, Liu, Chen, and Luo]{gpt4roi}
Shilong Zhang, Peize Sun, Shoufa Chen, Min Xiao, Wenqi Shao, Wenwei Zhang, Yu Liu, Kai Chen, and Ping Luo.
\newblock {GPT4RoI: Instruction Tuning Large Language Model on Region-of-Interest}.
\newblock \emph{arXiv preprint arXiv:2307.03601}, 2023{\natexlab{b}}.

\bibitem[Zhang et~al.(2023{\natexlab{c}})Zhang, Tian, Wu, Ji, Wang, Zhang, and Wan]{dvis}
Tao Zhang, Xingye Tian, Yu Wu, Shunping Ji, Xuebo Wang, Yuan Zhang, and Pengfei Wan.
\newblock {DVIS: Decoupled Video Instance Segmentation Framework}.
\newblock In \emph{Int. Conf. Comput. Vis.}, 2023{\natexlab{c}}.

\bibitem[Zheng et~al.(2024)Zheng, Qi, Chen, Wang, Wang, Qiao, and Zhao]{villa}
Rongkun Zheng, Lu Qi, Xi Chen, Yi Wang, Kun Wang, Yu Qiao, and Hengshuang Zhao.
\newblock {ViLLa: Video Reasoning Segmentation with Large Language Model}.
\newblock \emph{arXiv preprint arXiv:2407.14500}, 2024.

\bibitem[Zhou et~al.(2017)Zhou, Zhao, Puig, Fidler, Barriuso, and Torralba]{ade20k}
Bolei Zhou, Hang Zhao, Xavier Puig, Sanja Fidler, Adela Barriuso, and Antonio Torralba.
\newblock {Scene Parsing through ADE20K Dataset}.
\newblock In \emph{IEEE Conf. Comput. Vis. Pattern Recog.}, 2017.

\bibitem[Zhou et~al.(2022)Zhou, Wang, Zhang, Sun, Zhang, Birchfield, Guo, Kong, Wang, and Zhong]{avs}
Jinxing Zhou, Jianyuan Wang, Jiayi Zhang, Weixuan Sun, Jing Zhang, Stan Birchfield, Dan Guo, Lingpeng Kong, Meng Wang, and Yiran Zhong.
\newblock {Audio-Visual Segmentation}.
\newblock In \emph{Eur. Conf. Comput. Vis.}, 2022.

\bibitem[Zhu et~al.(2023)Zhu, Cheng, He, Li, Luo, Lu, Geng, and Xie]{trackgpt}
Jiawen Zhu, Zhi-Qi Cheng, Jun-Yan He, Chenyang Li, Bin Luo, Huchuan Lu, Yifeng Geng, and Xuansong Xie.
\newblock {Tracking with Human-Intent Reasoning}.
\newblock \emph{arXiv}, 2023.

\end{thebibliography}
}

\end{document}